\documentclass[12pt,oneside,english]{amsart}
\usepackage[T1]{fontenc}
\usepackage[utf8]{inputenc}
\usepackage{geometry}
\usepackage{fancyhdr}
\usepackage{color}
\usepackage{multirow}
\usepackage[table,xcdraw]{xcolor}
\usepackage{graphicx}
\usepackage[foot]{amsaddr}
\usepackage{amstext}
\usepackage{amsthm}
\usepackage{amssymb}
\usepackage{mathdots}
\usepackage{esint}
\usepackage{listings}
\usepackage{setspace}
\usepackage[noadjust]{cite}
\usepackage{caption, subcaption}
\usepackage[ruled]{algorithm2e}
\usepackage{soul}
\usepackage{comment}
\usepackage[normalem]{ulem}

\newcommand{\rulesep}{\unskip\ \vrule\ }

\makeatletter
\numberwithin{equation}{section}
\numberwithin{figure}{section}
\theoremstyle{plain}

  \theoremstyle{plain}
  
  \theoremstyle{remark}
  \newtheorem*{rem*}{\protect\remarkname}
  \theoremstyle{remark}
  
  \theoremstyle{plain}

\usepackage[english]{babel}
\usepackage{fancyhdr}
\definecolor{Code}{rgb}{0,0,0}
\definecolor{Decorators}{rgb}{0.5,0.5,0.5}
\definecolor{Numbers}{rgb}{0.5,0,0}
\definecolor{MatchingBrackets}{rgb}{0.25,0.5,0.5}
\definecolor{Keywords}{rgb}{0,0,1}
\definecolor{self}{rgb}{0,0,0}
\definecolor{Strings}{rgb}{0,0.63,0}
\definecolor{Comments}{rgb}{0,0.63,1}
\definecolor{Backquotes}{rgb}{0,0,0}
\definecolor{Classname}{rgb}{0,0,0}
\definecolor{FunctionName}{rgb}{0,0,0}
\definecolor{Operators}{rgb}{0,0,0}
\definecolor{Background}{rgb}{0.98,0.98,0.98}

\providecommand{\lemmaname}{Lemma}

\providecommand{\theoremname}{Theorem}
\providecommand{\corollaryname}{Corollary}
\newcommand{\abs}[1]{\ensuremath{|#1|}}

\newcommand{\cF}{\mathcal{F}}

\usepackage{babel}
\providecommand{\lemmaname}{Lemma}
  
  \providecommand{\remarkname}{Remark}
\providecommand{\theoremname}{Theorem}

\usepackage{babel}
\providecommand{\lemmaname}{Lemma}
  
\providecommand{\theoremname}{Theorem}

\usepackage{babel}
\providecommand{\theoremname}{Theorem}

\usepackage{babel}
\providecommand{\theoremname}{Theorem}

\usepackage{mathtools}

\DeclarePairedDelimiterX{\inp}[2]{\langle}{\rangle}{#1, #2}

\makeatother

\usepackage{babel}
  \providecommand{\corollaryname}{Corollary}
  \providecommand{\lemmaname}{Lemma}
  \providecommand{\remarkname}{Remark}
\providecommand{\theoremname}{Theorem}

\begin{document}

\title{Hedging using reinforcement learning:\\ Contextual $k$-Armed Bandit versus $Q$-learning}

\author{Loris Cannelli}
\author{Giuseppe Nuti}
\author{Marzio Sala}
\author{Oleg Szehr}

\address[Loris Cannelli, Oleg Szehr] {Dalle Molle Institute for Artificial Intelligence (IDSIA) - SUPSI/USI; Lugano; Switzerland}
\email{loris.cannelli@idsia.ch}
\email{oleg.szehr@idsia.ch}
\email{giuseppe.nuti@ubs.com}
\email{marzio.sala@ubs.com}

\address[Giuseppe Nuti, Marzio Sala]{{Quantitative Analyst; UBS Investment Bank; Zurich, New York}}

\thanks{This work is supported by UBS project LP-15403 / CW-202427\\{The views and opinions expressed in this material are those of the respective speakers and are not those of UBS AG, its subsidiaries or affiliate companies (``UBS''). Accordingly, neither UBS nor any of its directors, officers, employees or agents accepts any liability for any loss or damage arising out of the use of all or any part of this material or reliance upon any information contained herein.}}

\begin{abstract}
The construction of replication strategies for contingent claims in the presence of risk and market friction is a key problem of financial engineering. In real markets, continuous replication, such as in the model of Black, Scholes and Merton (BSM), is not only unrealistic but it is also undesirable due to high transaction costs. A variety of methods have been proposed to balance between effective replication and losses in the incomplete market setting. With the rise of Artificial Intelligence (AI), AI-based hedgers have attracted considerable interest, where particular attention was given to Recurrent Neural Network systems and variations of the $Q$-learning algorithm. From a practical point of view, sufficient samples for training such an AI can only be obtained from a simulator of the market environment. Yet if an agent was trained solely on simulated data, the run-time performance will primarily reflect the accuracy of the simulation, which leads to the classical problem of model choice and calibration. In this article, the hedging problem is viewed as an instance of a risk-averse contextual $k$-armed bandit problem, which is motivated by the simplicity and sample-efficiency of the architecture. This allows for realistic online model updates from real-world data. We find that the $k$-armed bandit model naturally fits to the Profit and Loss formulation of hedging, providing for a more accurate and sample efficient approach than $Q$-learning and reducing to the Black-Scholes model in the absence of transaction costs and risks.
\end{abstract}

\maketitle


\section{\label{subsec:itro}Introduction and Motivation}

The construction of dynamic hedging strategies lies at the core of the valuation and risk-management of derivative contracts in investment firms. The modern valuation theory came to light with two seminal articles of Black and Scholes \cite{black1973} and Merton \cite{merton1973theory} on the valuation of European options in an idealized economic model. This model, commonly known as the BSM economy, is characterized by continuous-time trading, a complete and frictionless market and stock price processes of geometric Brownian motion type. Black-Scholes and Merton construct a self-financing trading strategy in a riskless security (cash) and a risky stock, which perfectly replicates the option's payoff at maturity, thus \textit{hedging} the option's risk. The absence of arbitrage implies that the price equals the value of the replicating portfolio over the whole lifetime. In particular, the price of the option at inception is defined as the cost of setting up the replicating portfolio.

In reality, continuous time trading is, of course, impossible. It can not even serve as a reasonable approximation, due to the high resulting transaction costs. Instead, a \textit{replicating portfolio} is adjusted at discrete times, optimizing between replication error and trading costs. The acceptable deviation from the exact hedge depends on the risk tolerance of the investor. As a consequence, the best hedging strategy constitutes a trade-off between replication error (risk) and trading costs. To quantify this trade-off it is common to introduce a utility function, which reflects the investor's preference. The investor's hedging decisions are then guided by the maximization of expected utility over the investment horizon. Utility-based hedging entails a significant complication of the BSM approach: instead of minimizing the immediate replication error, the investor is faced with a multi-period planning problem, whose goal is the optimal trade-off between risk and global transaction costs. Furthermore, in the context of incomplete markets, it is far less immediate what the price of a derivative contract should be. This can possibly give rise to the bid-ask spread for the derivative contract, which tends to be extraordinarily wide in the absence of a portfolio based approach. In fact various concepts of \textit{price} occur in the literature, two common examples being the ``indifference'' and ``minimum-risk'' prices. Indifference purchase/sell prices are the amount of money that makes an investor indifferent, in terms of the expected utility, between trading in a market with and without purchasing/selling the derivative. The minimum-risk price is the amount of initial capital that maximizes an investor's utility given an optimal sequence of hedging decisions. Thus, pricing typically requires the solution of an exogenous optimization problem, given a hedging strategy, e.g.~to minimize the global risk. Utility-based hedging and reservation prices have been introduced in pioneering work by Hodges and Neuenberger~\cite{hodges1989option}. Minimum-risk prices were first studied in \cite{schweizer1995variance}. Hodges and Neuenberger also interpreted hedging in the language of dynamic programming and solved the respective Bellman equations, anticipating some of the modern developments.

With the rise of AI, an entire area of research emerged around the solution of dynamic programming problems, which gained wide public audience under the name of Reinforcement Learning (RL). Architectures that combine RL with deep Neural Networks (NNs) have proven super-human capabilities in various domains. The classical examples include video games of increasing complexity such as Atari~\cite{mnih2013playing} and Starcraft~\cite{vinyals2019grandmaster}, and strategic ($P$-) games like Hex~\cite{NIPS2017_d8e1344e}, Go and Chess~\cite{silver2016mastering,silver2017mastering}. Several publications have also reported on promising hedging results by RL agents, see e.g.~\cite{hodges1989option,kolm2019dynamic,halperin2017qlbs,cao2019deep}. In terms of the choice of algorithm,~\cite{kolm2019dynamic,halperin2017qlbs,cao2019deep} focus on variations of the $Q$-learning algorithm~\cite{watkins1992q} for the approximation of optimal policies in the hedging problem. The article \cite{halperin2017qlbs} proposes $Q$-learning as a tool for training hedging agents with a quadratic utility functional, but with no transaction costs. \cite{cao2019deep} applies double $Q$-learning \cite{van2016deep} to take account of stochastic volatility in the hedging problem. {Deep $Q$-Learning} (DQN)~\cite{osband2016deep} is a variation of $Q$-learning, which combines $Q$-learning with deep NNs to represent quality of state-action values. The articles~\cite{buehler2019deep,buehler2019deep2} take a slightly different approach to hedging, interpreting multi-period utility optimization as a {supervised learning} problem. In this setting a recurrent NN is trained via simultaneous optimization over an entire sequence of predictable hedging decisions. For pricing \cite{buehler2019deep,buehler2019deep2} follow the \textit{minimum-risk} approach, minimizing a given convex risk measure of the terminal holdings. The article~\cite{kolm2019dynamic} applies DQN to study a discrete-time hedging problem in the presence of transaction costs and served as the motivation for the investigation at hand. As compared to~\cite{hodges1989option,halperin2017qlbs,cao2019deep,buehler2019deep,buehler2019deep2} this article focuses on the hedging (rather than valuation) of option contracts assuming that price information is given in an exogenous form (e.g.~through a dedicated pricing engine). This assumption leads to what is called the \emph{accounting Profit and Loss} (P\&L) formulation of the hedging problem, see~\cite{cao2019deep}. In this formulation the investor's exercise is to balance between transaction costs and expected deviation from the price process. In contrast, in the \textit{Cash Flow formulation} the investor is only aware of the contractual payoffs of the derivative, which leads to a full-fledged RL problem.

All mentioned publications~\cite{hodges1989option,halperin2017qlbs,cao2019deep,buehler2019deep,buehler2019deep2, kolm2019dynamic} assume an accurate market simulator to train their AI agents. However, setting up such a simulator leads to the classical financial engineering problem of model choice and calibration. This might be seen as a bottleneck for applying those hedging systems in practice as the run-time performance of the AI will often merely reflect the accuracy of the simulator. In this sense it can also hardly be claimed
that the trained agents operate in a fully model-free setup. This would apply if enough
market data was available for training, which is rarely the case in
realistic derivative markets. In reality, run-time performance will be driven by the accuracy of the market simulator
- making an AI-based approach awfully similar to standard Monte Carlo methods in use for decades -
unless the agent can quickly adapt to unseen scenarios. At this point sample efficiency is key, as the comparatively
small amount of real-world data that the agent encounters in operation will otherwise not influence the training in a statistically significant way. Similarly, an agent trained on simulated data will usually not acquire the {{knowledge}} to appropriately capture market impact in a practical setting. The latter would require training on samples with market impact, which will often be hard to provide in sufficient amount. In the case of a simulated market impact, such samples will simply present the modeled cost of the market impact. The main use-case for current AI systems can thus be seen in hedging of complex derivatives or large derivative portfolios that cannot be efficiently handled using conventional pricing engines as in~\cite{buehler2019deep,buehler2019deep2}.

The article at hand argues that - in a practical setting - a bandit-type RL algorithm can be used to address hedging in the P\&L formulation. Our theoretical considerations are accompanied with extensive computer experiments comparing the hedging performance of our NN bandit algorithm versus DQN. The $k$-armed bandit is the prototypical instance of RL and has been studied extensively within diverse areas such as {advertising} selection, clinical trials, finance, etc., see~\cite{sutton2018reinforcement} for examples. As compared to Supervised Learning, the bandit algorithm adapts to new market data \textit{online}, i.e.~in the course of its execution, by collecting information about realized utility/reward signals. As compared to full-fledged RL, in particular, to variants of $Q$-learning, the bandit is characterized by the assumption that rewards at different time steps are independent and identically distributed. As a consequence each action only determines the immediate reward but has no influence on the future. In terms of hedging, this implies that the agent optimizes for an immediate utility signal and the agent's choices do not dynamically impact the market. As compared to DQN, bandit algorithms are simpler, better studied, possess stronger performance guarantees, are easier to train, and results can be interpreted more immediately. Most importantly, significantly less samples are required during the training process with well-known optimal regret estimates~\cite{auer2002finite}. In summary the perspective taken in the article at hand is as follows:
\begin{enumerate}
\item[(i)] Standard microscopic market models employed to train AI agents do not capture any dependency on trading decisions of market participants and strong market impact signifies a singular situation.
\item[(ii)] Any AI trained on {simulated data} acquires a behavior typical of the simulation and it will perform well mostly in situations that are similar to the simulation. In particular, it will not be able to make informed decisions in an unusual market state.
\item[(iii)] If the number of available training samples is limited, the restricted planning capability of the $k$-armed bandit as compared to more sophisticated RL systems is more than out-weighted by its higher sample efficiency.
\end{enumerate}
The article is structured as follows. Section~\ref{bmsection} describes the methodological background for this article, introducing utility-based hedging in Section~\ref{hedg} and relevant background from RL in Sections~\ref{cmabsection} and \ref{q_intro}. Section~\ref{sec_algos} provides technical details of the algorithms chosen for our experiments. Specifically, Section~\ref{cmab} describes the NN-based contextual bandit and Section~\ref{dql} the deep $Q$-learning algorithm. Our experimental findings are presented in Section~\ref{sec:proofofc}. Section~\ref{sec:conclude} concludes the presentation.

\section{Background and Methodology}
\label{bmsection}

\subsection{Hedging and Pricing}\label{hedg}

We first describe the full dynamic-programming approach to hedging in incomplete markets in the Cash Flow formulation and specialize to the P\&L formulation later. Although we focus on option hedging the discussion extends similarly to more general contracts, see~\cite{duffie2001dynamic} for details. We assume that an investor (agent) sells an European option and maintains a hedging portfolio to off-set the risk until maturity $T>0$. The agent receives new market information at a series of discrete times $t = 1,2,.... <T$, re-balancing the hedging portfolio immediately once the new information becomes available. It is assumed throughout that the random market behavior follows a Markov process, which does not depend on the choice of hedges. Assume the hedging portfolio is composed of $n_t$ shares of stock $S_t$, and $B_t$ units of cash
and let $\Pi_t$ denote its value at time $t$:
\begin{equation*}\Pi_t=n_tS_t+B_t.\end{equation*}
For shorter notation we assume that there is no compounding on the cash holdings, since this can be incorporated in straight-forward way into our framework. Once the new market information is available the agent chooses the number $n_{t+1}$ of shares for the coming period. While the agent adjusts the number of shares, any purchase or sell is equally billed to the bank account $B_t$, which corresponds to a \emph{self-financing} trading strategy. In the presence of transaction costs each trading activity is accompanied with an additional loss: $cost<0$, and the trading constraint becomes:
\begin{align}
&\Delta\Pi_{t+1}=n_{t+1}S_{t+1}+\Delta B_{t+1}-n_tS_t + cost(n_t,n_{t+1},S_t)\\
&=n_{t+1}\Delta S_{t+1} + cost(n_t, n_{t+1}, S_t)\nonumber.
\end{align}
Here and in what follows we adopt the convention $\Delta X_{t+1} := X_{t+1}-X_t$. The agent's goal is to choose hedging decisions $(n_1,n_2,...,n_T)$ as to maximize the expected utility of terminal wealth $w_T$:
\begin{align}
\max\,\mathbb{E}[u(w_T)],\label{toBeMaximized}
\end{align}
where the utility function $u(\cdot)$ is smooth, increasing and concave. The terminal wealth $w_T$ is the sum of the replicating portfolio $\Pi_T$ and the payoff $C_T$ of the option:
\begin{equation*}C_T=-(S_T-K)^+.\end{equation*}
The dynamic-programming view on the hedging problem is as follows: one time step before maturity the agent chooses action $n_T$ such as to maximize the conditional expectation:
\begin{align*}
\mathbb{E}[u(C_T+\Pi_T) |\cF_{T-1}].
\end{align*}
Conditioning on the filtration $\cF_{T-1}$ means that all market information at time $T-1$, including $S_{T-1}$, $\Pi_{T-1},...,$ has been realized and is available for decision making. The agent proceeds by planning backwards in time. Two steps before maturity the agent chooses action $n_{T-1}$ to maximize the above expectation conditioned on $\cF_{T-2}$, three steps before maturity $n_{T-2}$ to maximize conditioned on $\cF_{T-3}$, and so on. The optimal course of action corresponds to the optimal \emph{value function}:
$$V^*(t,S_t,\Pi_t) = \text{max}_{n_1,n_2,...}\mathbb{E}[u(C_T+\Pi_T) |\cF_{t}].$$
It is a classical result that this function can be obtained as a solution of the Bellman optimality equation~\cite{puterman1994markov}. RL is concerned with the computation of approximate solutions to this equation. Remark that the successive optimization of utility expectations corresponds to the Cash Flow formulation for the hedging problem.

We illustrate the dynamic programming perspective on the hedging problem by an example. Consider a frictionless market and let the terminal wealth be split into an initial endowment and a series of wealth increments:
$$w_T = w_0+\sum_t \Delta w_{t+1}.$$
In the absence of transaction costs the wealth increment is $\Delta w_{t+1} =\Delta C_{t+1}+\Delta \Pi_{t+1}$.
Suppose an agent maximizes~\eqref{toBeMaximized} by minimizing a risk measure of individual increments:
\begin{equation*}
Risk_t=\mathbb{E}\left[(\Delta w_{t+1})^2|\mathcal{F}_t\right].
\end{equation*}
This corresponds to a time-additive utility of the form $u(w_T)=\sum_t u_t(\Delta w_{t+1})$ with quadratic $u_t$.
The agent then proceeds by choosing $n_T$ to minimize $Risk_{T-1}$, then $n_{T-1}$ to minimize $Risk_{T-2}$, etc.. At each time step the price $C_t$ can be defined as the minimizer of $Risk_t$ given the optimal hedge. In the limit of a continuous action space and continuous time this procedure gives rise to the familiar BSM hedging. The optimal action is the $\delta$-hedge:
\begin{equation*}
\delta_t=\frac{\partial C^*_t}{\partial S_t},
\end{equation*}
where $C^*_t$ denotes the BSM option price, leading to a complete off-set of risk and a utility of $0$. In the presence of transaction costs and discretization error, risk-free hedging is not possible. The computation of optimal hedges requires the solution of the full-fledged RL problem, because transaction costs introduce reward-dependencies even for a time-additive utility. Standard pricing concepts, such as the mentioned indifference and minimum risk prices, rely on the optimal course of hedging decisions and the solution of the multi-period problem as an input.

The situation is simplified drastically in the P\&L formulation, where prices $C_t$ are available a priori throughout the planning process. In this situation the agent is merely faced with an optimization between transaction costs and the immediate risk. Assuming transaction costs and a time-additive utility, a possible choice is the minimization of the immediate risk of the form:
\begin{equation}
Risk_t=\mathbb{E}\left[(\Delta C_{t+1} + n_{t+1}\Delta S_{t+1}+cost(n_t, n_{t+1},S_t))^2|\mathcal{F}_t\right].\label{formulaBSD_tk}
\end{equation}
It should be mentioned that if $C_t$ are computed externally by a pricing engine without accounting for transaction-costs, one might be tempted to interpret equations of the form \eqref{formulaBSD_tk} (or~\eqref{formulaBSD_tk_MeanVar} below) as a ``way to generalize hedging'' to the situation of transaction costs and discretization error. In theory, computing hedges in this way, is not guaranteed to result in the optimal terminal utility, even approximately. The definition of prices via risk-minimization or indifference pricing requires the full solution of the RL problem with transaction costs, while the optimization given pre-computed prices will, in general, not provide a meaningful approximation. However, and this is one of the main messages of the article at hand, in the practical work of an investment firm, the course of hedging decisions is dictated rather by the availability of information than the seek for theoretically optimal pricing. In the absence of a market model the exact computation of indifference/minimum-risk prices becomes elusive, leaving the investor with the task of choosing~\emph{reasonable} hedges given available data and infrastructure. In the sequel we follow the line of~\cite{kolm2019dynamic} focusing on the accounting P\&L formulation, which provides an efficient approximate hedging framework in practice. In what follows we argue that, both from a theoretical and practical perspective, if RL is applied to this problem, then a contextual bandit-type algorithm should be employed.

\subsection{Contextual Multi-armed Bandit (CMAB) Model}\label{cmabsection}
The $k$-armed bandit is the prototypical instance of RL and has been studied extensively within various areas of practical interest~\cite{sutton2018reinforcement}. The setup involves a set of $k\in\mathbb{N}_+$ possible choices\footnote{The name originates from a gambler who chooses from $k$ slot machines.} and a sequence of $T$ periods. In each period $t = 1,...,T$ the learner chooses action $a\in\{a_1,..,a_k\}$ and receives a random reward $r_t(a) = r(a,t)$. The main assumption is that rewards are independent over $a$ and i.i.d.~over $t$. The objective is to identify a ``policy'', i.e.~a probabilistic rule of action, in order to maximize {the} cumulative expected reward:
\begin{equation*}\sum\limits_{t=1}^T\mathbb{E}[r_t],\end{equation*}
over the execution episode. In the context of derivatives hedging this form of reward-optimization corresponds to a time-additive utility functional in~\eqref{toBeMaximized}, where each $r_t$ stands for the utility of a wealth increment. In the \textit{contextual} $k$-armed bandit setting, the agent is faced with non-stationary reward distributions. In each round, prior to taking its decision, the agent receives {\textit{context}} about the state of the system. Rewards then depend on the context in a hidden way, which has to be learned by the agent. For example, an agent could be presented with the tuple $(t, n_t, C_t, S_t)$ to assist the maximization of~\eqref{formulaBSD_tk}. The contextual bandit involves both the association of rewards with the given context but also trial-and-error learning to search for strong actions. Contextual search tasks are {an} intermediate between the prototypical $k$-armed bandit and the full RL problem~\cite{sutton2018reinforcement}. They are like the full RL problem in that they involve learning a policy, but each action affects only the immediate reward. Like any RL agent, the contextual $k$-armed bandit must bargain between {exploration and exploitation}: is it better to choose a lucrative action given a context or should the agent explore in hope to find something even better? A priori exploration implies risk in the sense that the agent must deviate from an optimal action. To take account of the hedging risk {we employ} a \emph{risk-averse} version of CMAB, which we call R-CMAB. 

While in the standard bandit problem, the objective is to select the arm associated to the highest reward, the mean-variance bandit aims at most effectively trading off expected rewards versus variance (i.e.~risk). The mean-variance of action $a$ is defined as:
\begin{align}
MV(a):=\kappa\mu(a)-\sigma^2(a),\label{meanVar}
\end{align}
where $\kappa\in[0,1]$ measures the risk aversion, and $\mu(\cdot)$ and $\sigma(\cdot)$ are, respectively, the mean and the standard deviation associated to action $a$. The optimization of the mean-variance under transaction costs for hedging has been proposed in~\cite{kolm2019dynamic} and constitutes a generalization of the reward structure~\eqref{formulaBSD_tk}. For learning the agent builds up an empirical reward statistics. For sample outcomes $\{X_l(a)\}_{l=1,...,T}$ of action $a$, the empirical mean-variance at a given time $t$ is:
\begin{align*}
\widehat{MV}_t(a)=\kappa\hat{\mu}_t(a)-\hat{\sigma}_t(a),
\end{align*}
with empirical mean and standard deviation given by:
\begin{align}
\hat{\mu}_t(a)=\frac{1}{t}\sum_{l=1}^tX_l(a),\quad\textnormal{and}\quad\hat{\sigma}_t(a)=\frac{1}{t}\sum_{l=1}^t(X_l(a)-\hat{\mu}_t(a))^2.
\end{align}
The mean-variance multi-armed bandit problem has been introduced in~\cite{sani2012risk}, where also the Mean-Variance Lower Confidence Bound algorithm is proposed. Notice also that the risk-averse multi-armed bandit algorithm has performance and convergence guarantees, see~\cite{sani2012risk,vakili2016risk} for details. To our knowledge the article at hand contains the first application of a mean-variance contextual bandit system.

\subsection{$Q$-learning}
\label{q_intro}
$Q$-learning is a classical algorithm for the off-policy approximation of solutions to general RL problems beyond the maximization of immediate rewards~\cite{watkins1992q}.
The agent stores a $Q$-table that contains {estimates} of the $Q$-uality of state-actions pairs $(s_t,a_t)$ in view of their terminal reward\footnote{Remark that the state $s_t$ at time $t$ can contain multiple items of information, e.g.~$s_t=(t,n_t,S_t,C_t)$ for option hedging.}. At each time step, the agent \emph{i)} chooses the best action $a_t$ from the table, \emph{ii)} observes the corresponding reward $r_t$, and \emph{iii)} updates the estimates of the $Q$-table according to a specific updating rule. For $Q$-learning this rule is given by:
\begin{equation}
\hat{q}_{new}(s_t,a_t)\leftarrow \hat{q}_{old}(s_t,a_t)+\alpha\left[r_{t+1}+\gamma\underset{a}{\max}\:\hat{q}_{old}(s_{t+1},a)-\hat{q}_{old}(s_t,a_t)\right],\label{Qlearn}
\end{equation}
where $\hat{q}$ denotes the empirical estimate of $Q$ values. A key point is that being a temporal-difference method, $Q$-learning performs the update~\eqref{Qlearn} immediately after each time step as opposed to plain Monte Carlo methods that execute:
\begin{equation*}
\hat{q}_{new}(s_t,a_t)\leftarrow \hat{q}_{old}(s_t,a_t)+\alpha\left[\sum_{t=1}^Tr_{t}-\hat{q}_{old}(s_t,a_t)\right],
\end{equation*}
after each learning episode is completed. Thus~\eqref{Qlearn} contains two types of estimates. First, there is the estimation of $q(s_t,a_t)$ via a Monte Carlo estimator $\hat{q}_{new/old}(s_t,a_t)$. As for an ordinary Monte Carlo method, the step-size parameter $\alpha\in(0,1]$ in~\eqref{Qlearn} weights the relevance of the new sample versus the held estimate $\hat{q}_{old}$. Second, the quantity $r_{t+1}+\gamma\underset{a}{\max}\: \hat{q}(s_{t+1},a)$ serves as a Bellman estimate of the total expected reward. $\gamma\in(0,1]$ is a discount factor that reduces the value of rewards that are farther in the future. The iterative application of this procedure is guaranteed to converge to the optimal policy if the step-size is small enough~\cite{sutton2018reinforcement}.
%
%
%
DQN algorithms enhance this procedure by making use of a (deep) NN for the representation of $Q$ values. The NN is trained on Monte Carlo samples once sufficient data is available, boosting the quality of the samples of subsequent simulations, see Section~\ref{dql} for details.

\subsection{Machine Learning Algorithms for Financial Derivatives Hedging}
$Q$-learning addresses the computation of optimal hedges in the Cash Flow formulation. In principle it can be employed in situations of strong environment feedback, reflecting the impact of trading decisions to the market. Implementations of $Q$-learning for the Cash Flow formulation can be found in~\cite{halperin2017qlbs,cao2019deep} and for the P\&L formulation in~\cite{kolm2019dynamic}. For hedging in the P\&L formulation it is sufficient to optimize the immediate rewards, which can be achieved by contextual bandit-type algorithms. For the P\&L formulation we follow~\cite{kolm2019dynamic} in choosing the mean-variance reward function (extending~\eqref{formulaBSD_tk}):
\begin{align}
u_t=&\kappa\mathbb{E}\left[(\Delta C_{t+1} + n_{t+1}\Delta S_{t+1}+cost(n_t, n_{t+1},S_t))|\mathcal{F}_t\right]\nonumber\\
&-\mathbb{V}\left[(\Delta C_{t+1} + n_{t+1}\Delta S_{t+1}+cost(n_t, n_{t+1}, S_t))|\mathcal{F}_t\right]\label{formulaBSD_tk_MeanVar},
\end{align}
where $\mathbb{V}[X]$ stands for the variance of $X$. This is of the form~\eqref{meanVar} and fits with the reward structure of the R-CMAB algorithm.

RL algorithms are capable of learning without human supervision and experience. This comes at the cost of a larger computational effort and less sample efficiency as compared to problem-tailored methods and supervised learning. 
In the derivative hedging area, the articles~\cite{buehler2019deep,buehler2019deep2} interpret the optimization problem~\eqref{toBeMaximized} as a supervised learning exercise. The expected utility is maximized by an optimization over the whole predictable sequence of hedging decisions $(n_1,n_2,...,n_T)$ by training a recurrent NN on random market paths. In sophisticated domains like the games of Chess or Go the gradient descent-based optimization of rewards will usually get stuck at a poor local maximum because of the high complexity of the value function or non-differentiable reward structure. However, in hedging scenarios, where a market model is available, gradient descent-based optimization provides a sample-efficient and stable approach to the hedging problem, which also scales favorably in multi-asset portfolio optimization tasks~\cite{buehler2019deep,buehler2019deep2}.
In recent years variations of $Q$-learning have became somewhat outdated for planning problems, when an explicit model (such as the rules of a game, or a market) is available. Neural Monte Carlo Tree Search (MCTS) algorithms learn stronger policies faster and can be seen as the next step of technological evolution, see e.g.~\cite{silver2017mastering,NIPS2017_d8e1344e}. MCTS combines RL with search techniques to direct the learning process to promising courses of actions. As compared to plain Monte Carlo methods (including $Q$-learning),
MCTS builds a problem-specific, asymmetric and heavily restricted Monte Carlo decision tree. The performance of the famous Atari DQN~\cite{mnih2013playing} has, for instance, been improved using MCTS~\cite{guo2014deep}. Similar to~\cite{buehler2019deep,buehler2019deep2}, MCTS easily learns the hedging of financial derivatives in the Cash Flow formulation~\cite{szehr2021hedging} in case that a market model is given a priori. In case no such model is available, the exact solution of the hedging and pricing problems in the Cash Flow formulation will often be elusive. The only viable option seems to revert to a bandit model and P\&L hedging as a practical approximation. This is illustrated by our experimental findings, Section~\ref{sec:proofofc}, which give a detailed comparison of our R-CMAB algorithm to DQN. A heuristic comparison of machine learning algorithms, which is based on figures presented in the mentioned publications and our findings, is given in Table~\ref{sumtable}. The market model column shows if a market model is required in principle, although in realistic operations all models will require a market simulator due to lack of data.
\begin{table}
\includegraphics[scale=0.46]{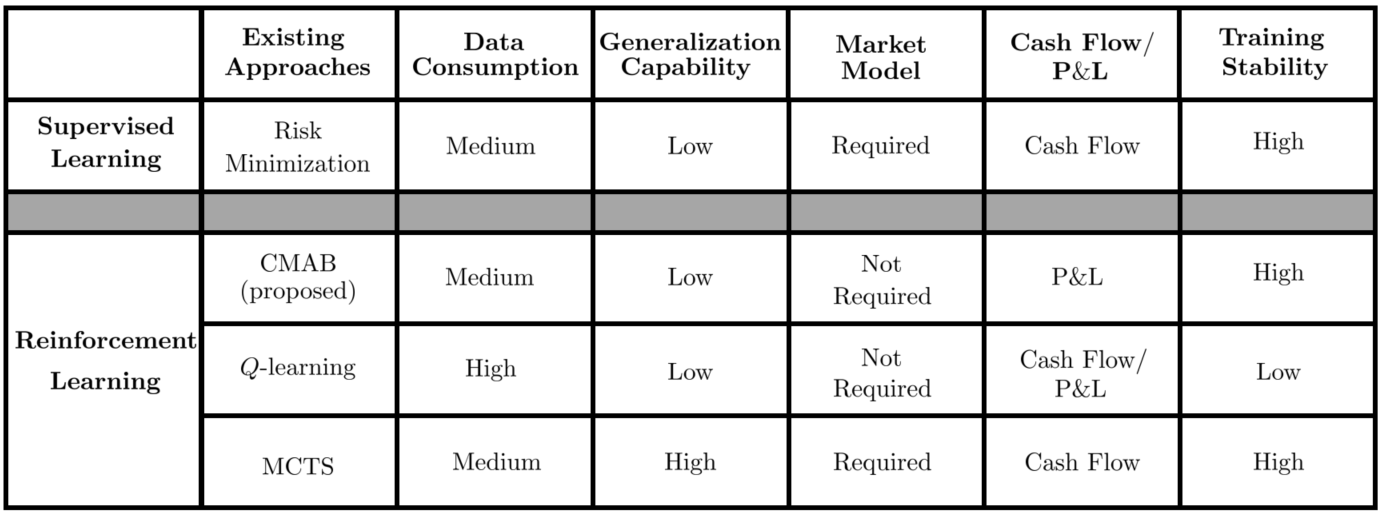}
\caption{Heuristic comparison of some features of algorithms for derivatives hedging in the literature. }\label{sumtable}
\end{table}

\section{Algorithms}\label{sec_algos}

This section summarizes the chosen architectures for the contextual bandit and $Q$-learning algorithms, which are well-known in the literature. In general RL algorithms operate by executing episodes of actions and considering the obtained rewards in order to improve subsequent courses of action. In order to achieve high rewards the algorithm needs to exploit the reward-information acquired from previous episodes; in order to improve performance it needs to explore potentially sub-optimal courses of action. Broadly, the various RL algorithms differ in the ways how the acquired information is represented and how they handle the trade-off between exploration and exploitation. To achieve a good performance, appropriate representation power and an intelligent coordination of explore/exploit decisions over the learning process are key. In the implementation at hand these points are addressed by the usage of NNs to link contexts and rewards and by making use of Bayesian posterior sampling to coordinate the courses of action. Bayesian posterior sampling maintains a Bayesian estimate of a distribution over reward models through the entire learning process. As reward observations are gathered, this distribution is updated according to Bayes’ rule, sharpening it around the expected rewards of each action. At each decision step a set of parameters is sampled and the best action is taken according to the resulting model. As compared to just taking the action of highest estimated reward, this sampling procedure ensures a sufficient level of exploration, where the concentration of the posterior around the expected reward reflects the increasing confidence in the reward estimate. The entire procedure leans at Thompson sampling~\cite{thompson1933likelihood,chapelle2011empirical} in the non-contextual case but can be viewed as a method to approximate non-tractable Bayesian updates also in the presence of a complex context model. Since market moves are random, the RL agents must operate in an environment of high stochasticity. While hedging has been explored with $Q$-learning, the application of bandit-type algorithms in such an environment seems new. We have experimented with different architectures, including purely linear inference, variational inference and plain Monte Carlo, similar to~\cite{riquelme2018deep}, where the chosen design has shown the best performance. All experiments were conducted using PyTorch libraries on an ordinary notebook, Intel(R) Core(TM) i7-8565U CPU @ 1.80GHz and 16Gig of RAM.

\subsection{Neural Network R-CMAB algorithm}\label{cmab}

To achieve high representation power for the contextual bandit the association of contexts to rewards is reflected by a NN. A direct way to coordinate exploration is to assume a linear Bayesian regression model on top of the last hidden layer of the NN as in~\cite{riquelme2018deep,snoek2015scalable}. While the NN is trained to accurately predict rewards, a sampling procedure replaces the deterministic output layer. Concretely, the predicted value $v_a$ for action $a$ is given by $v_a = \beta_a^T z_x$. Here $z_x$ denotes the output of the last hidden layer given context $x$ and the parameter vectors $\beta_a$ are sampled from the posterior corresponding to action $a$. The posterior at step $t$ is computed following
$\pi_t(\beta_a,\sigma_a^2) = \pi_t(\beta_a|\sigma_a^2) \pi_t(\sigma_a^2),$
where $\sigma_a^2$ reflects our confidence about the reward from $a$ and we assume $\sigma_a^2\sim IG$, and $\beta_a|\sigma_a^2\sim N$ with an Inverse Gamma and Normal distribution, respectively. The exact rules for the parameter updates are given in~\cite{riquelme2018deep,snoek2015scalable}. The correlations present in the sequence of contexts might cause instability of the training process as the NN tends to overfit to this correlation. As a common method to remedy this instability we employ a memory buffer, which stores the transitions $(s_t,a_t,r_t,s_{t+1})$ for a given time step. During the training process random batches are sampled from this buffer, breaking the inter-temporal correlations. The neural network and the posterior model are updated asynchronously at given updating frequencies. The neural R-CMAB algorithm is summarized in Algorithm \ref{algo1}. 

\begin{algorithm}[h]
    \SetAlgoLined
	\textbf{Input} Init.~action-value NN with random parameters; Set up prior distribution over models, $\pi_0:\theta\in\Theta\rightarrow[0,1]$. Init.~replay memory to given capacity\;
	\For{episode $=1:numberOfEpisodes$}{
    	\For{$t=1:length(episode)$}{
    		Sample parameters $\theta_t\sim\pi_t$\;
    		Get a context vector $x_t\in\mathbb{R}^d$\;
    		Compute $a_t=\text{BestAction}(x_t,\theta_t)$ through NN followed by linear Bayesian model\;
    		Perform action $a_t$ on hedging environment and observe reward $r_t$\;
            Add $(x_t,a_t,r_t)$ to memory buffer\;
    		\If{updateModelFrequencyCriterium}{
    		    Update Bayesian estimate of the posterior distribution to $\pi_{t+1}$ using most recent transitions\;}
    		\If{updateNNFrequencyCriterium}{
    		    Sample random batches from memory buffer to train NN\;}
    	}
	}
	\caption{Neural R-CMAB Algorithm}
	\label{algo1}
\end{algorithm}

\subsection{Neural Network $Q$-learning}\label{dql}
The DQN algorithm has been originally designed for learning from {alternating visual} data \cite{mnih2013playing}. Our architecture orients itself broadly at the Atari game example of~\cite{mnih2013playing} and the R-CMAB above. Similarly to the CMAB algorithm, the $Q$-values are represented internally by a NN, which is trained on approximate reward targets and exploration is managed as in~\cite{mnih2013playing} by adding random noise with decaying probability. Following the temporal-difference paradigm the parameters of the NN are updated intra-episode as in~\eqref{Qlearn} on Bellman targets of the form $r_{t+1}+\gamma\underset{a}{\max}\: \hat{q}(s_{t+1},a)$. The gradient then points into the direction of $r_{t+1}+\gamma\underset{a}{\max}\:\hat{q}_{old}(s_{t+1},a)-\hat{q}_{old}(s_t,a_t)$ as in~\eqref{Qlearn}, where $\alpha$ takes the role of the learning rate. As opposed to supervised learning, where learning targets are fixed, the Bellman targets depend on the NN parameters themselves, which would lead to training the NN on a sequence of targets that change at each learning iteration. Two instances of the same NN equipped with different suits of parameters are employed to stabilize the learning process. The policy NN provides the estimates $\hat{q}_{old}$ and is trained at each iteration on the Bellman targets. The target NN provides the action values for the computation of Bellman targets. The parameters of the target NN are updated asynchronously every $C$ iteration of the learning process with the parameters of the policy NN. As for the CMAB we employ a replay memory buffer, which stores the transitions $(s_t,a_t,r_t,s_{t+1})$ for a given time step. During the training of the policy NN random batches are sampled from this buffer. The NN is then updated by the batch stochastic gradient descent algorithm, which serves to stabilize the update via~\eqref{Qlearn}. The described DQN algorithm is summarized in Algorithm~\ref{algo2}. Finally, let us mention that many variations of the DQN design have appeared in the literature, where more recent architectures are known to outperform the original design in terms of stability and sample efficiency, see e.g.~\cite{hessel2018rainbow}.

\begin{algorithm}[H]
	\SetAlgoLined
	\textbf{Input} Init.~action-value target and policy NNs with random parameters; Set up exploration parameter: $\epsilon$. Init.~replay memory to given capacity\;
	\For{episode $=1:numberOfEpisodes$}{
		\For{$t=1:length(episode)$}{
			Get current state of environment $s_t\in\mathbb{R}^d$\;
			Compute action $a_t=\text{BestAction}(s_t,\theta_t)$ through policy NN followed by $\epsilon$-greedy action\;
			Perform $a_t$ on hedging env.~and observe reward $r_t$, new state $s_{t+1}$\;
			Store transition $(s_t,a_t,r_t,s_{t+1})$ in the replay memory\;
			Sample mini-batch of transitions from the replay memory\;
			Compute Bellman targets for states in this mini-batch by evaluating target NN\;
			Update the policy NN parameters by one step of stochastic batch gradient descent on Bellman targets\;
			\If{updateModelFrequencyCriterium}{
		    Update exploration parameter $\epsilon$\;}
			}
		\If{updateTargetNNFrequencyCriterium}{
		    Update parameters of target NN by policy NN\;}
	}
	\caption{Neural (Deep) $Q$-learning Algorithm}
	\label{algo2}
\end{algorithm}

\section{Proof of concept and experimental findings}\label{sec:proofofc}
This section compares the performance of R-CMAB and DQN in view of a potential deployment in a setting, where {training data is limited}. We focus on the P\&L formulation, which means that the hedger has access to price information through an exogenous pricing engine and we compute rewards via~\eqref{formulaBSD_tk_MeanVar}. We focus on the hedging of a short European option contract in a discrete-time setting, but we expect similar outcomes for more general derivative contracts. As is standard in RL the training process is organized in independent ``training episodes'', see~\cite{sutton1988learning} for an introduction. An episode corresponds to the full life-time of the option contract from inception to maturity. All relevant option and market data is reset to the same values at the beginning of each episode. Each episode is split into a number of equal time intervals, where hedging decisions occur in the beginning of each interval. Training proceeds by considering ``reward-samples'' (i.e. market-feedback) that the agent receives as a consequence of the preceding hedging decision over multiple episodes. In what follows we will often start the training process ``tabula rasa'', i.e., the agent possesses no prior information but learns solely from the observed reward statistics. Usually we will not attempt to train an agent ``to perfection'' on simulated data, but rather we are interested in the performance after a comparatively small number of training episodes. We summarize the basic settings for our comparison.
\begin{itemize}
\item \textbf{Option contract:} Short Euro vanilla call with strike $K=1.0$ and maturity $T=60$ days.
\item \textbf{Market:} Underlying stock price is given by a Geometric Brownian Motion, $S_0=1.0$, with $\mu=0.01$, $\sigma=0.3$. Flat interest rate market.
\item \textbf{Hedging Agent:} Hedging proceeds in discrete time. Each training episode (corresponding to $T=60$ days) is discretized into a number $10,20$ or $60$ time steps (samples).
\item \textbf{Action Space:} Actions are gauged to the interval $[-1,1]$ if short selling is allowed and to $[0,1]$ if not. We consider $51$ actions in the former case and $26$ actions in the latter.
\end{itemize}
The following setting has been chosen for the algorithms:
\begin{itemize}
\item\textbf{R-CMAB algorithm:} NN composed of $3$ layers with $20$ fully connected nodes each. ReLu activation. SGD optimizer with initial learning rate of $0.001$. Context is given by $(S_t,T-t)$. Reward function as in~\eqref{formulaBSD_tk_MeanVar} with scalar gauging by a factor of $100$ and a reward cut-off at $2$.

\item\textbf{DQN algorithm:} NN composed of $3$ layers with $20$ fully connected nodes each. ReLu activation. SGD optimizer with initial learning rate of $0.001$. State is given by $(S_t,T-t)$. Reward function as in~\eqref{formulaBSD_tk_MeanVar} with scalar gauging by a factor of $100$ and a reward cut-off at $2$.
\end{itemize}
Sections~\ref{trainingAlgos} and~\ref{comparisonAlgos} compare the training and test performance. For this a small discount factor of $\gamma=0.5$ has been chosen in DQN. Typical discount factors found in the literature fall in the range $0.99$, chosen in~\cite{mnih2013playing}, to $0.999$ \cite{lin1992reinforcement}. The effect of a small value for $\gamma$ is that DQN prioritizes the optimization of immediate rewards over planning for potential future rewards. In other words smaller $\gamma$ brings DQN closer to a contextual bandit algorithm, which creates favorable training conditions in the P\&L formulation of the hedging problem, see~\eqref{formulaBSD_tk_MeanVar}. Our choice gives DQN an artificial advantage for the comparison with our CMAB algorithm. However, we stress that this choice defies the purpose of a multi-period algorithm in the first place. For this reason we ran the experiments on pre-training and transaction costs in Sections~\ref{preTraining} and~\ref{transactionCosts} with the more realistic value $\gamma=0.9$.
Unsurprisingly the training speed of DQN deteriorates as $\gamma$ is increased towards higher values, which is illustrated in Section~\ref{comparisonAlgos} and discussed in Section~\ref{hedgingInPractice}. In addition we grant DQN a two times larger training set for all experiments.

\subsection{The training process}\label{trainingAlgos}
\begin{figure}[!htb]
	\centering
	\begin{subfigure}{0.48\linewidth}
		\includegraphics[width=\linewidth]{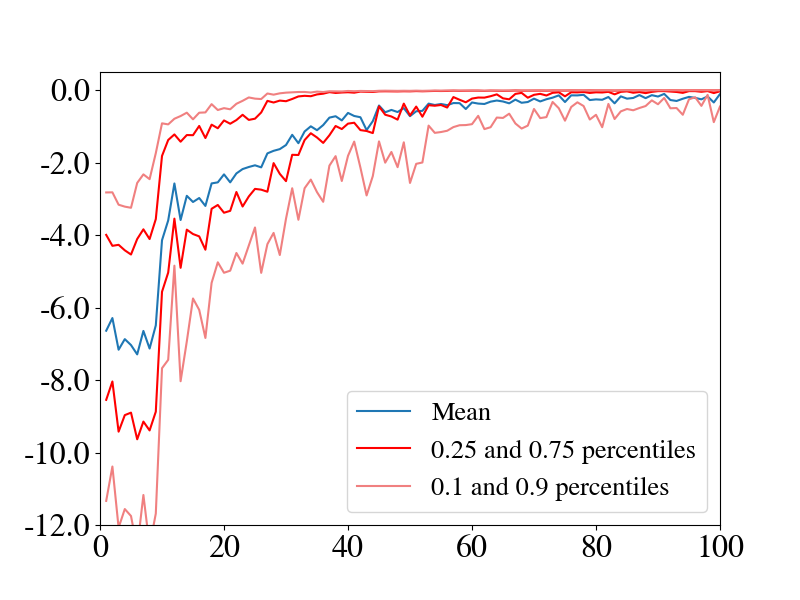}
		\caption{CMAB - 10 samples per episode}
	\end{subfigure}
	\hfil
	\rulesep
	\begin{subfigure}{0.48\linewidth}
		\includegraphics[width=\linewidth]{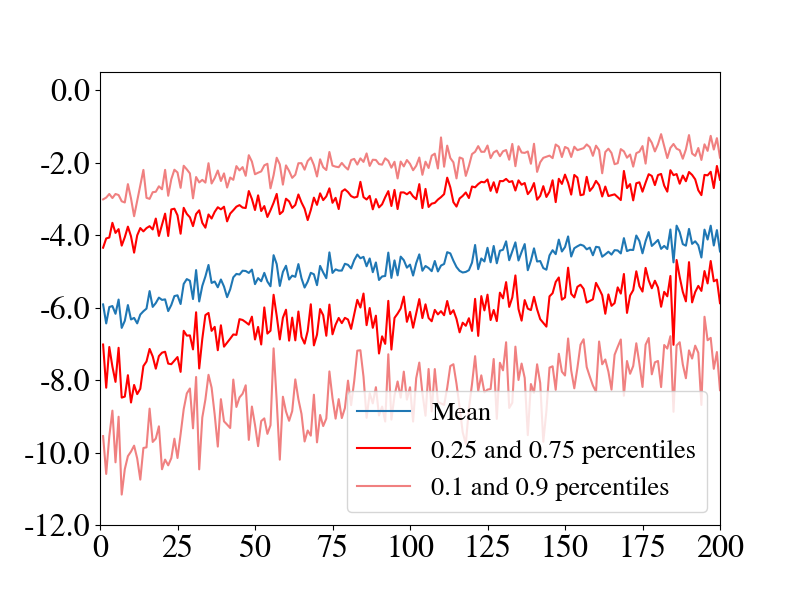}
		\caption{DQN - 10 samples per episode}
	\end{subfigure}
	
	\begin{subfigure}{0.48\linewidth}
		\includegraphics[width=\linewidth]{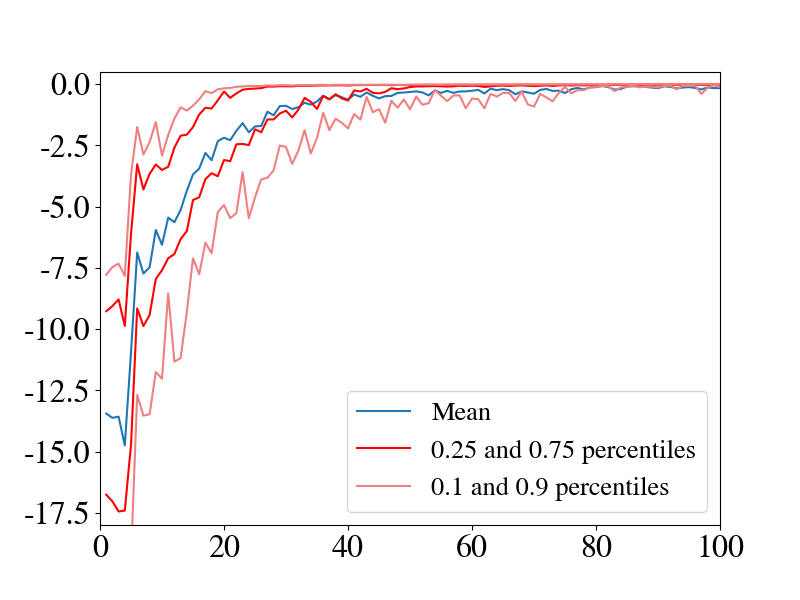}
		\caption{CMAB - 20 samples per episode}
	\end{subfigure}
	\hfil
	\rulesep
	\begin{subfigure}{0.48\linewidth}
		\includegraphics[width=\linewidth]{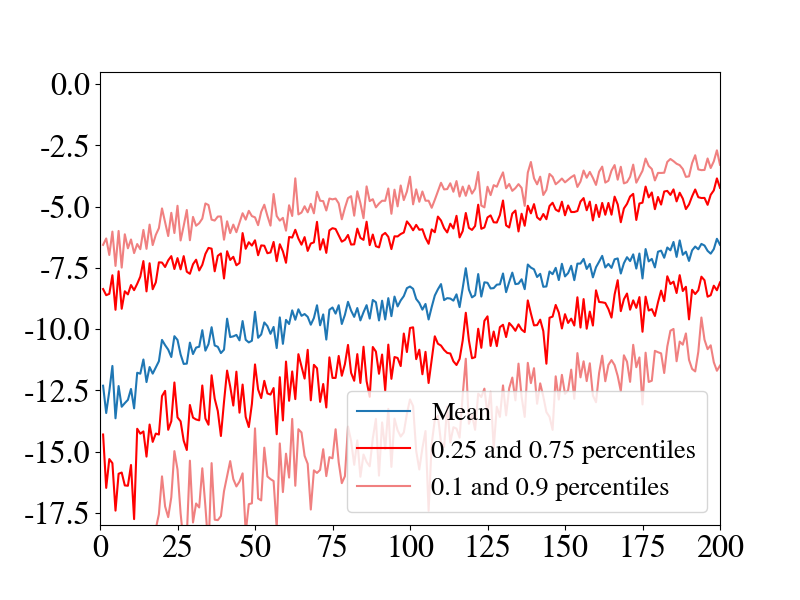}
		\caption{DQN - 20 samples per episode}
	\end{subfigure}
	
	\begin{subfigure}{0.48\linewidth}
		\includegraphics[width=\linewidth]{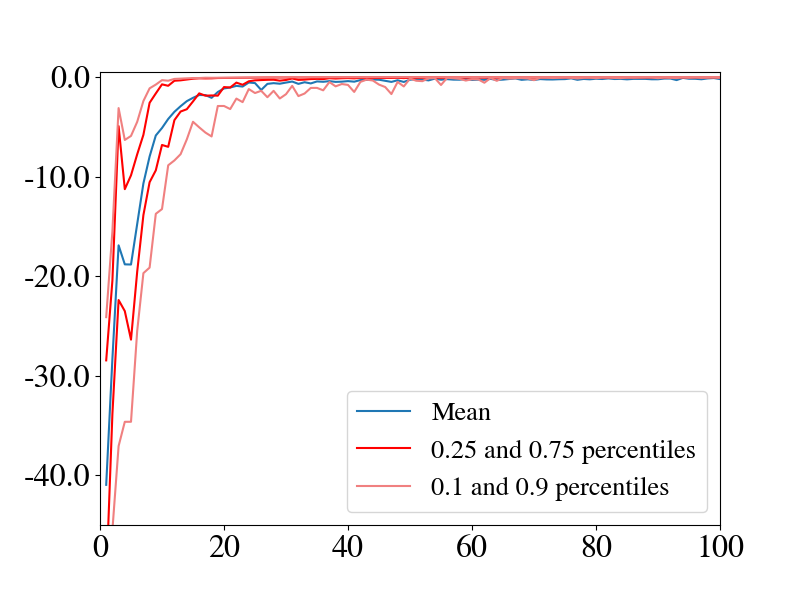}
		\caption{CMAB - 60 samples per episode}
	\end{subfigure}
	\hfil
	\rulesep
	\begin{subfigure}{0.48\linewidth}
		\includegraphics[width=\linewidth]{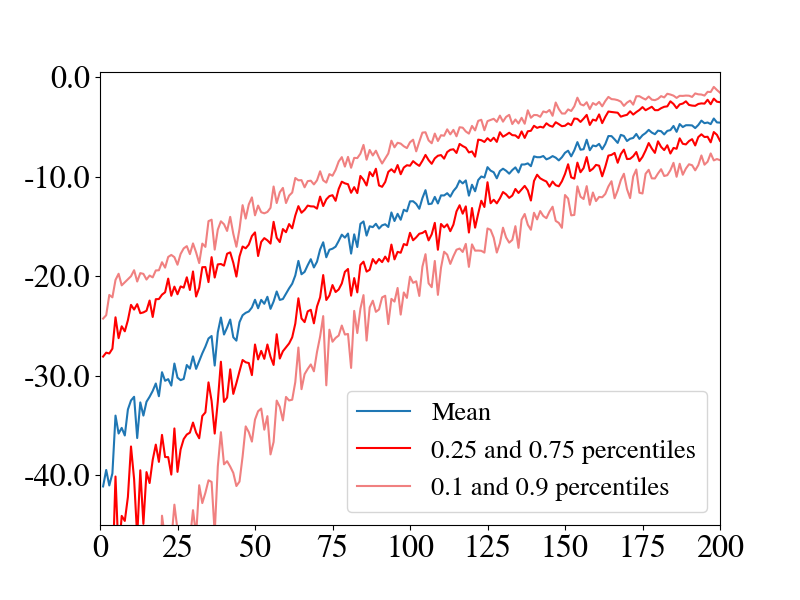}
		\caption{DQN - 60 samples per episode}
	\end{subfigure}
	\caption{Comparison of training progress of CMAB, $\kappa=0$, (left column) and DQN (right column) in the deterministic scenario. Graphs show descriptive statistics of episodic rewards obtained from $100$ independent training cycles.}
	\label{deterministic}
\end{figure}

We compare the training progress of CMAB and DQN algorithms by presenting some descriptive statistics from purely variance-based hedging (i.e.~$\kappa=0$ in~\eqref{formulaBSD_tk_MeanVar}). For this experiment, at each time step the agent chooses from $51$ possible hedging decisions, corresponding to $25$ buy/sell actions and one hold. (Short-selling is explicitly allowed here, although we are hedging a short call option.) To gain a basic quantitative picture we begin with a situation, where we pass the exact expectation value~\eqref{formulaBSD_tk_MeanVar} as a reward after each time step. In what follows we call this the ``deterministic scenario''. Notice that the computation of this expectation requires information about the underlying market model, which often will not be available in practice. Figure~\ref{deterministic} presents a comparison of episodic rewards during the training process in the deterministic situation. We ran $100$ independent training cycles of CMAB (with $100$ training episodes) and DQN (with $200$ training episodes) for this comparison.

\begin{figure}[!htb]
	\centering
	\begin{subfigure}{0.48\linewidth}
		\includegraphics[width=\linewidth]{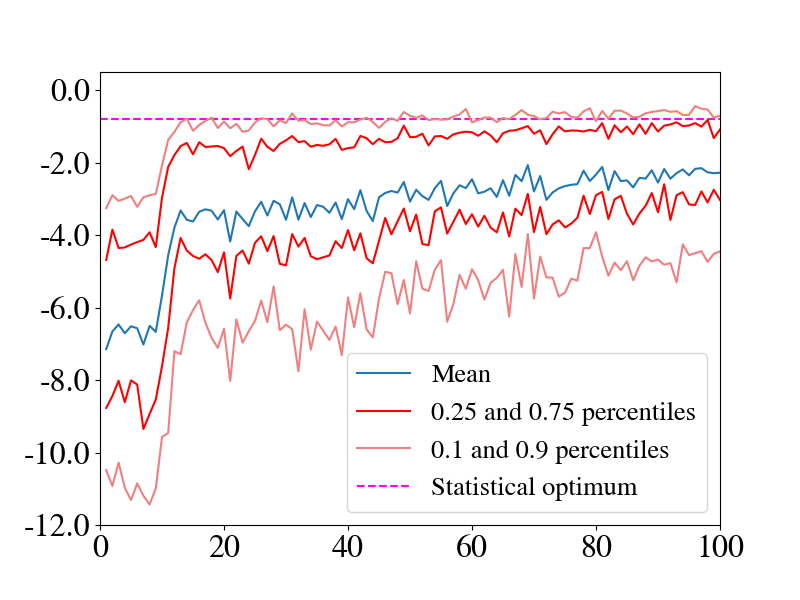}
		\caption{CMAB - 10 samples per episode}
	\end{subfigure}
	\hfil
	\rulesep
	\begin{subfigure}{0.48\linewidth}
		\includegraphics[width=\linewidth]{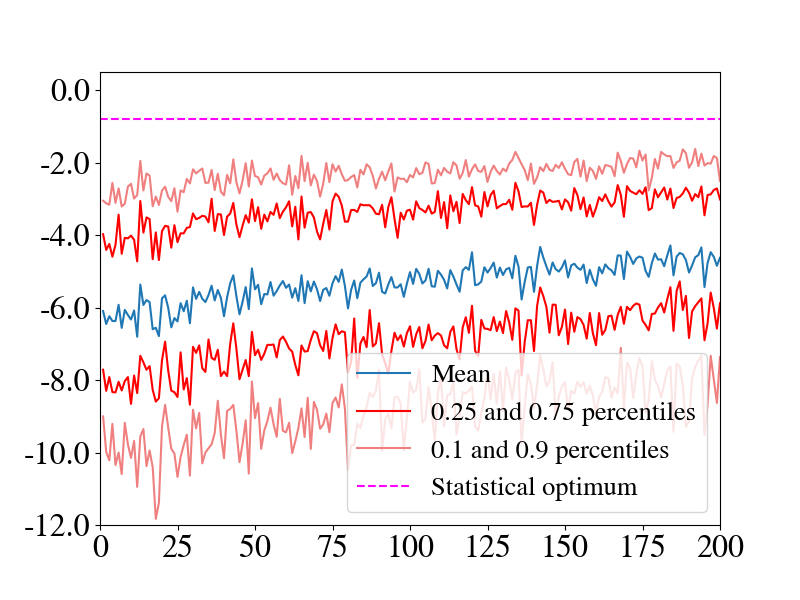}
		\caption{DQN - 10 samples per episode}
	\end{subfigure}
	
	\begin{subfigure}{0.48\linewidth}
		\includegraphics[width=\linewidth]{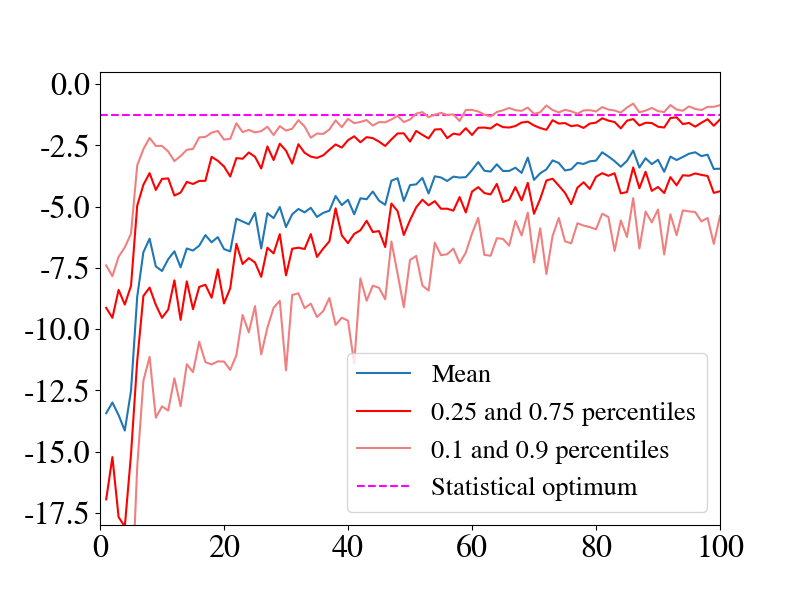}
		\caption{CMAB - 20 samples per episode}
	\end{subfigure}
	\hfil
	\rulesep
	\begin{subfigure}{0.48\linewidth}
		\includegraphics[width=\linewidth]{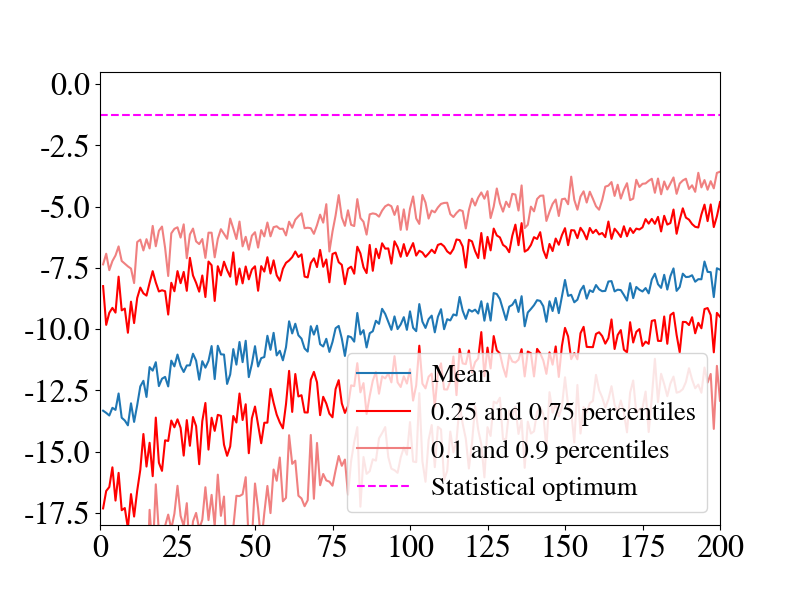}
		\caption{DQN - 20 samples per episode}
	\end{subfigure}
	
	\begin{subfigure}{0.48\linewidth}
		\includegraphics[width=\linewidth]{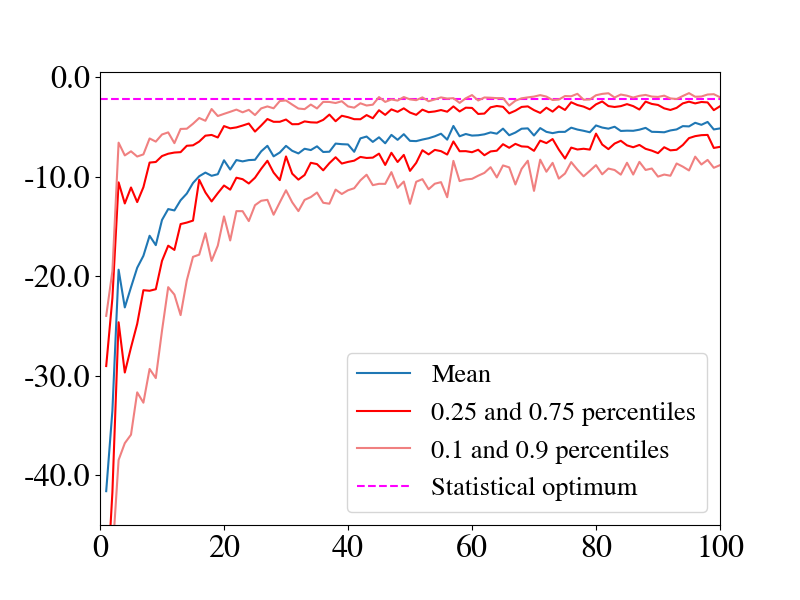}
		\caption{CMAB - 60 samples per episode}
	\end{subfigure}
	\hfil
	\rulesep
	\begin{subfigure}{0.48\linewidth}
		\includegraphics[width=\linewidth]{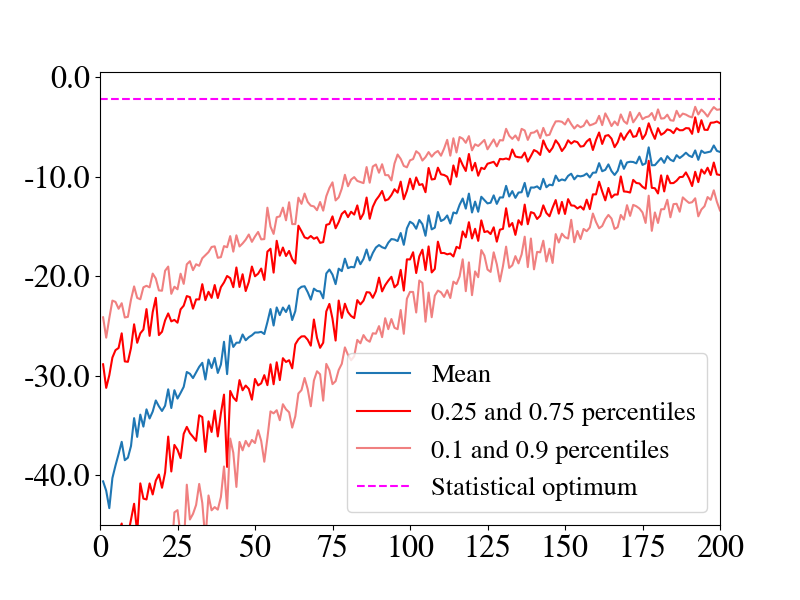}
		\caption{DQN - 60 samples per episode}
	\end{subfigure}
	\caption{Comparison of training progress of CMAB, $\kappa=0$, (left column) and DQN (right column) in the non-deterministic scenario. Graphs show descriptive statistics of episodic rewards obtained from $100$ independent training cycles.}
	\label{MDP}
\end{figure}

In the absence of a market model, the expectation~\eqref{formulaBSD_tk_MeanVar} is unknown and the agent learns from the random rewards obtained during the training process. In this situation the agent is left with choosing hedges and building up an internal model of the market's hidden reward structure according to the realized rewards. We call this the ``non-deterministic scenario''. The training progress in the non-deterministic situation is presented in Figure~\ref{MDP}. As for the deterministic case we ran $100$ independent training cycles of CMAB (with $100$ training episodes) and DQN (with $200$ training episodes) for this comparison. Since the agent decides on a hedge first and the reward is realized only one time-step later, this setting comes with an inevitable residual risk (here ``residual variance'' since $\kappa=0$) that cannot be hedged away. In other words, as compared to the familiar BSM economy, time- and action-space discretization entail a minimum rate of hedging error in the setting at hand. We computed this risk for an optimal oracle' hedger, that knows the expectation~\eqref{formulaBSD_tk_MeanVar} and hedges accordingly, and depict it by horizontal lines in Figure~\ref{MDP}.
\newpage{}
\newpage{}
\subsection{Comparison after training}\label{comparisonAlgos}
\begin{figure}[!htb]
	\centering
	\begin{subfigure}{0.9\linewidth}
		\includegraphics[width=\linewidth]{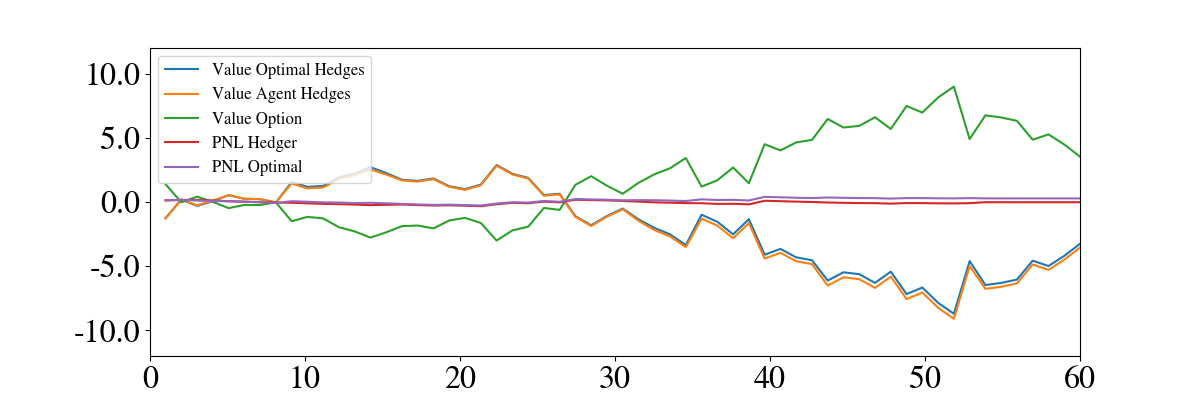}
		\caption{CMAB - $100$ episodes, $60$ samples per episode, deterministic scenario}
	\end{subfigure}
	\begin{subfigure}{0.9\linewidth}
		\includegraphics[width=\linewidth]{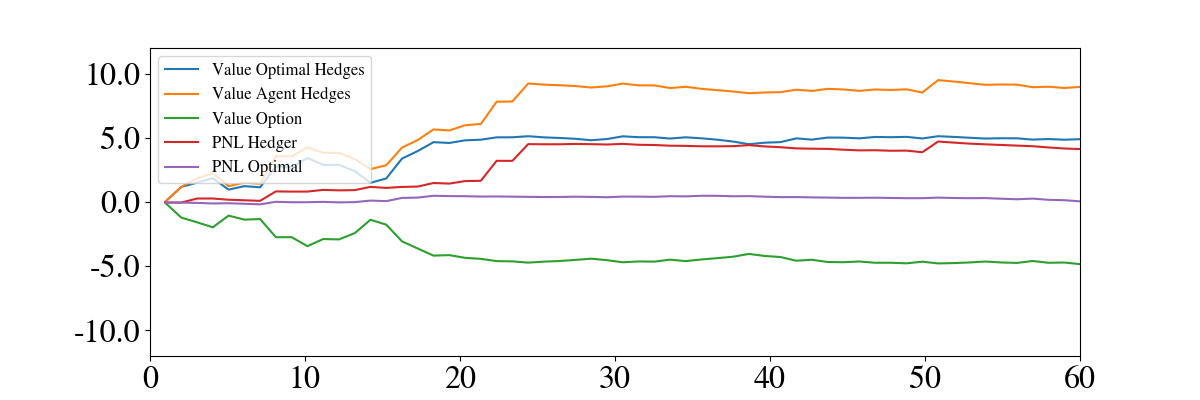}
		\caption{DQN - $200$ episodes, $60$ samples per episode, deterministic scenario}
	\end{subfigure}
	\begin{subfigure}{0.9\linewidth}
		\includegraphics[width=\linewidth]{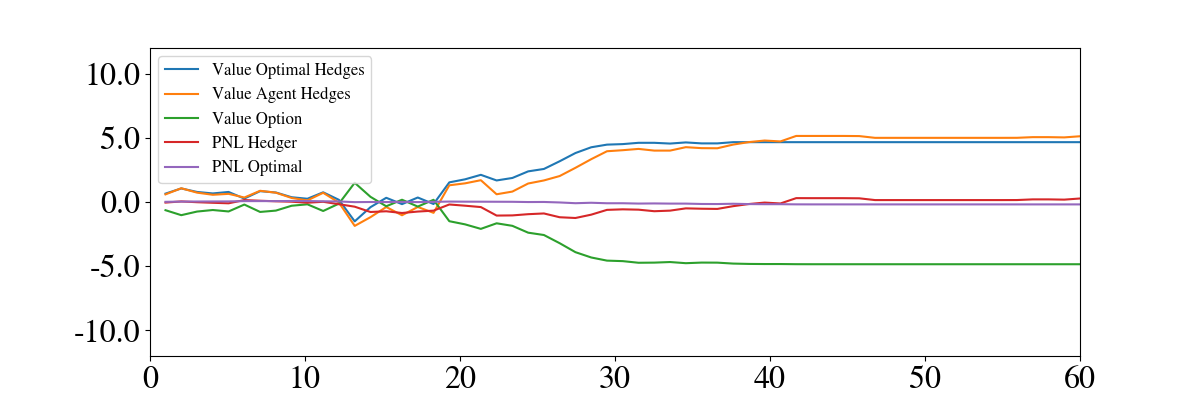}
		\caption{CMAB - $100$ episodes, $60$ samples per episode, non-deterministic scenario}
	\end{subfigure}
	\begin{subfigure}{0.9\linewidth}
		\includegraphics[width=\linewidth]{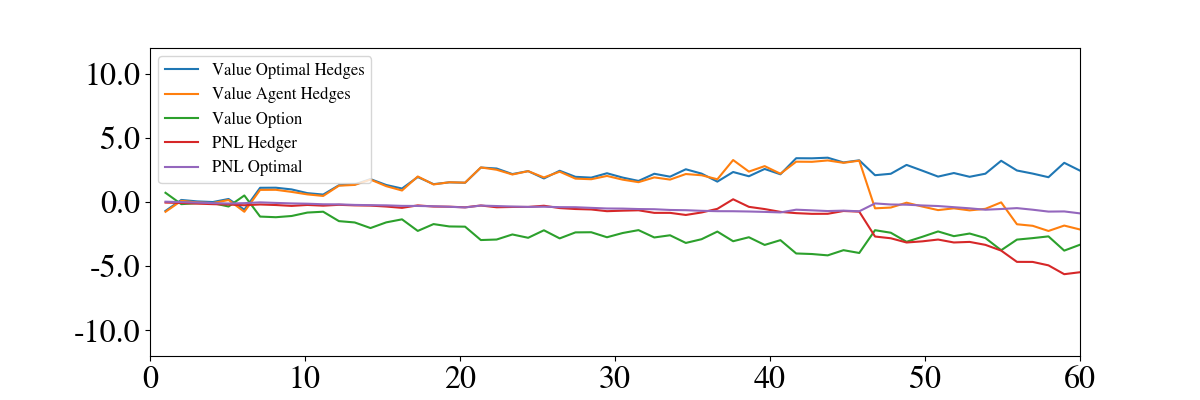}
		\caption{DQN - $200$ episodes, $60$ samples per episode, non-deterministic scenario}
	\end{subfigure}
	\caption{Exemplary hedging behavior of trained CMAB and DQN hedgers.}
	\label{examplePaths}
\end{figure}

\begin{figure}[!htb]
	\centering
	\begin{subfigure}{0.45\linewidth}
		\includegraphics[width=\linewidth]{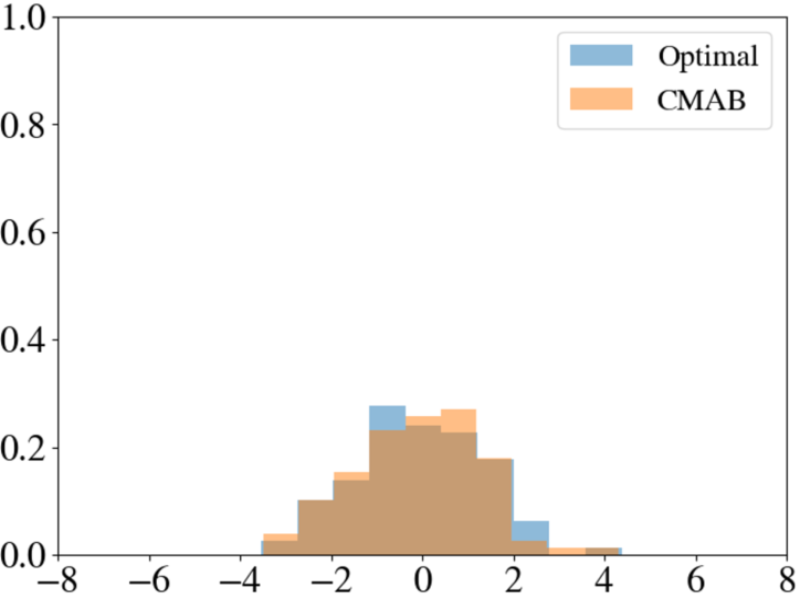}
	    \caption{CMAB - 10 samples per episode}
	\end{subfigure}
	\hfil
	\rulesep
	\begin{subfigure}{0.45\linewidth}
		\includegraphics[width=\linewidth]{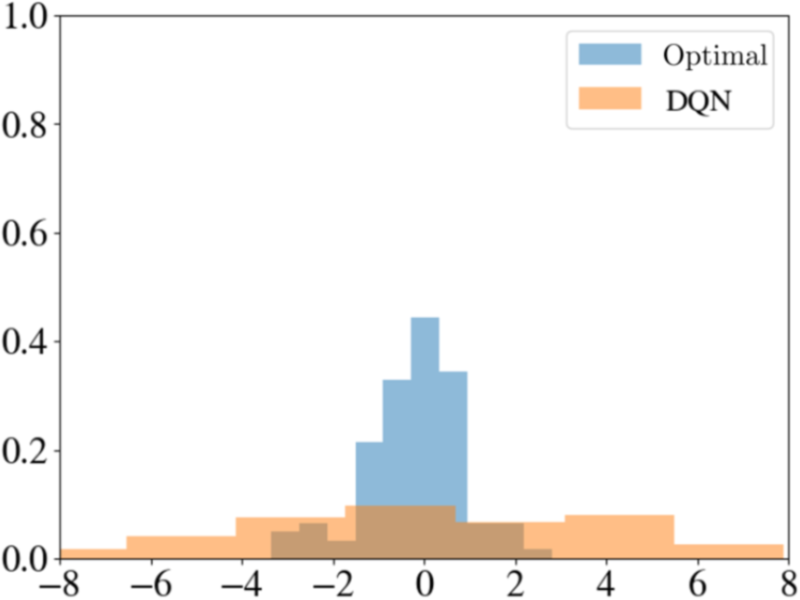}
	    \caption{DQN - 10 samples per episode}
	\end{subfigure}
	\begin{subfigure}{0.45\linewidth}
		\includegraphics[width=\linewidth]{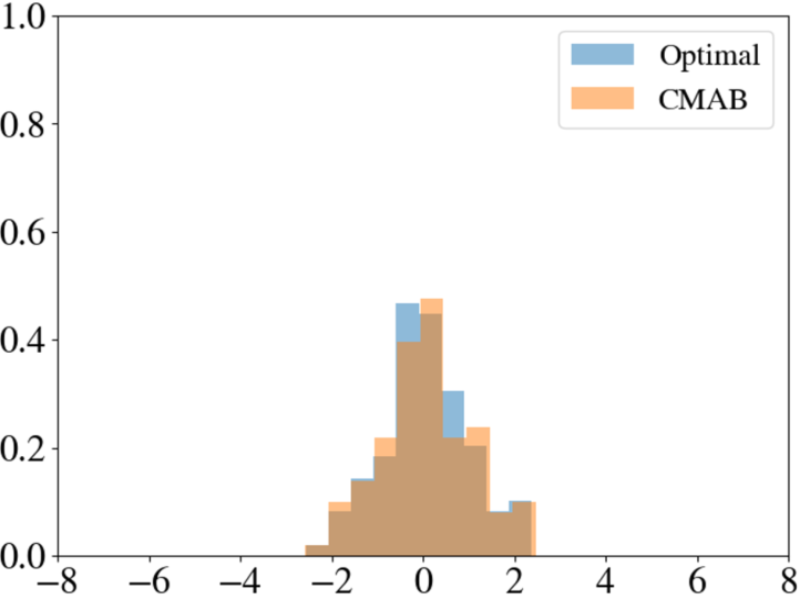}
		\caption{CMAB - 20 samples per episode}
	\end{subfigure}
	\hfil
	\rulesep
	\begin{subfigure}{0.45\linewidth}
		\includegraphics[width=\linewidth]{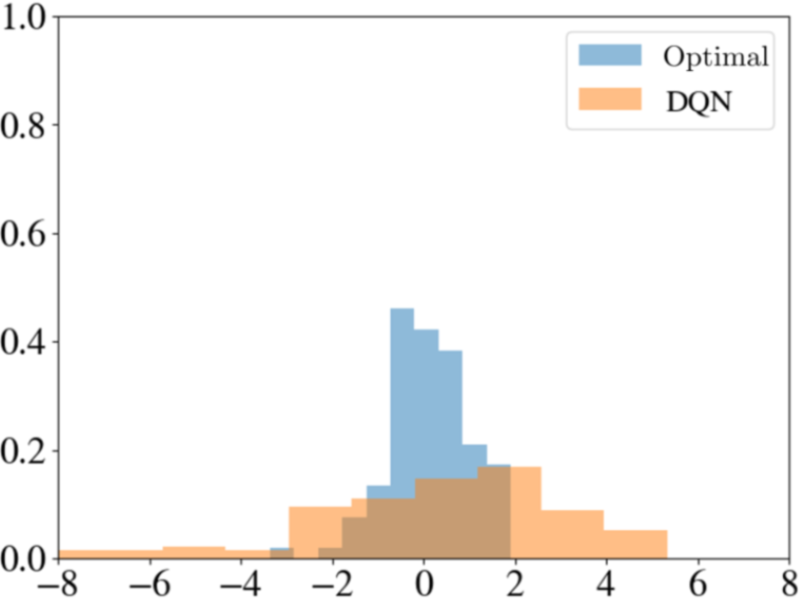}
		\caption{DQN - 20 samples per episode}
	\end{subfigure}
	
	\begin{subfigure}{0.45\linewidth}
		\includegraphics[width=\linewidth]{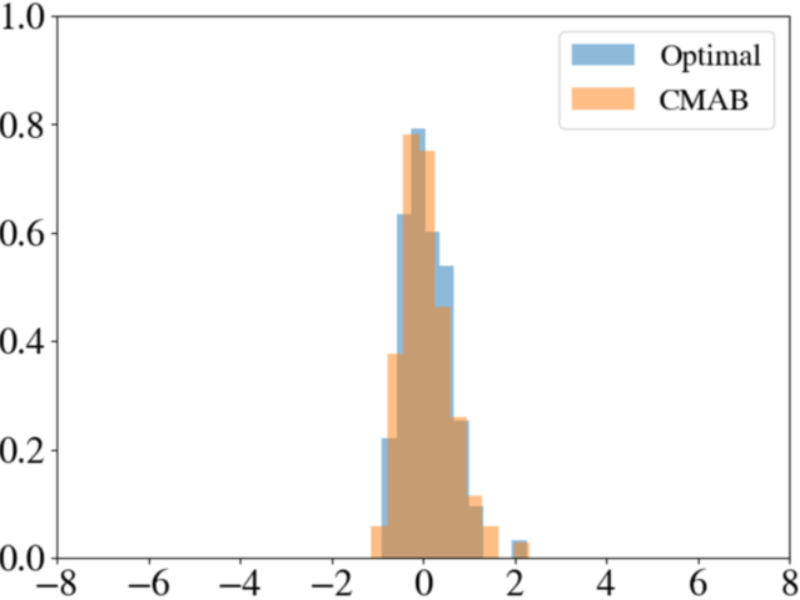}
		\caption{CMAB - 60 samples per episode}
	\end{subfigure}
	\hfil
	\rulesep
	\begin{subfigure}{0.45\linewidth}
		\includegraphics[width=\linewidth]{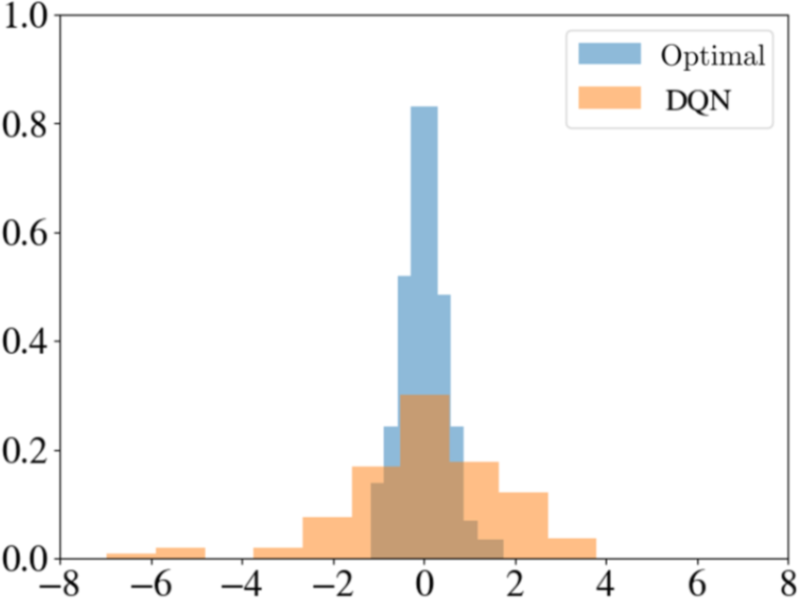}
		\caption{DQN - 60 samples per episode}
	\end{subfigure}
	\caption{Histograms of the terminal P\&L measured on $100$ test paths of CMAB, $\kappa=0$, (left column) and DQN (right column) agents in the deterministic scenario.}
	\label{plHists}
\end{figure}

\begin{figure}[!htb]
	\centering
	\begin{subfigure}{0.45\linewidth}
		\includegraphics[width=\linewidth]{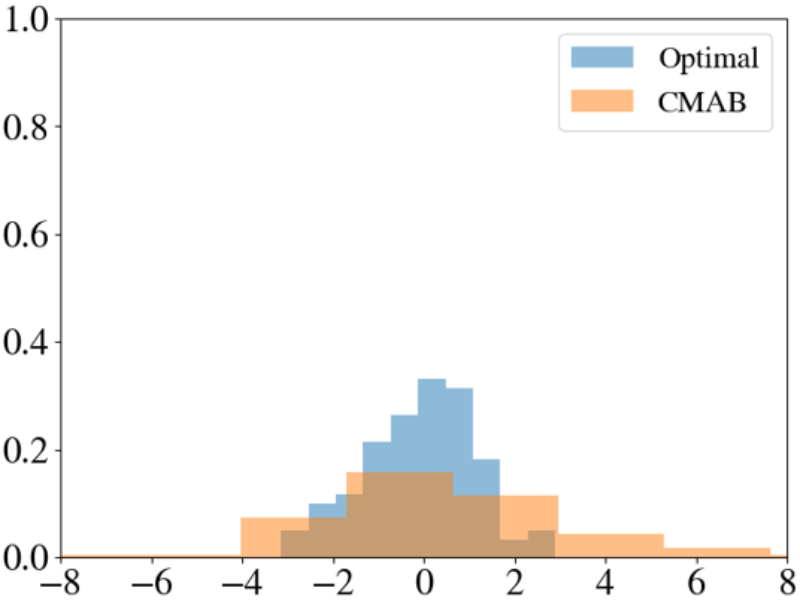}
		\caption{CMAB - 10 samples per episode}
	\end{subfigure}
	\hfil
	\rulesep
	\begin{subfigure}{0.45\linewidth}
		\includegraphics[width=\linewidth]{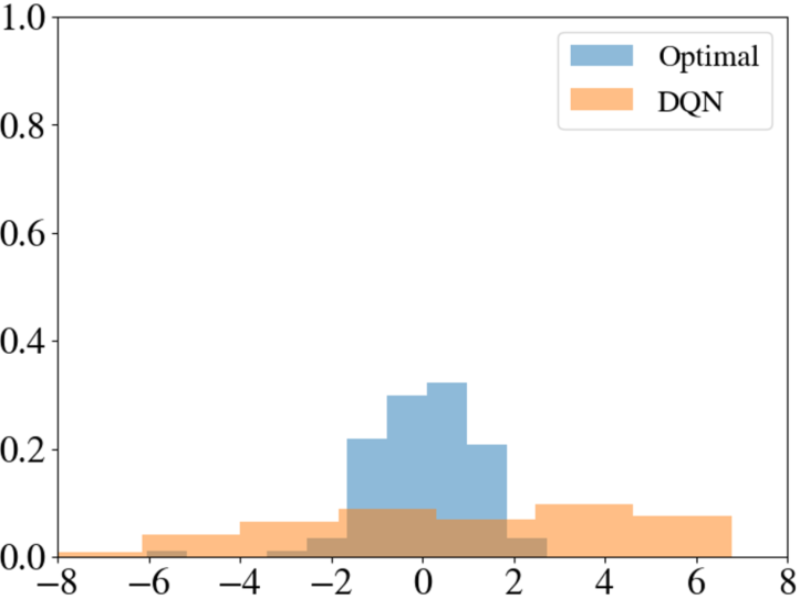}
		\caption{DQN - 10 samples per episode}
	\end{subfigure}
	\begin{subfigure}{0.45\linewidth}
		\includegraphics[width=\linewidth]{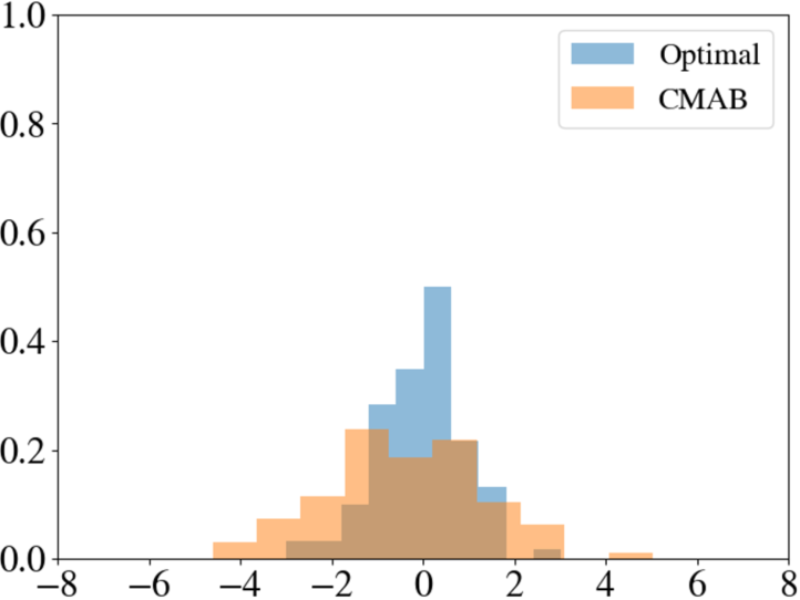}
		\caption{CMAB - 20 samples per episode}
	\end{subfigure}
	\hfil
	\rulesep
	\begin{subfigure}{0.45\linewidth}
		\includegraphics[width=\linewidth]{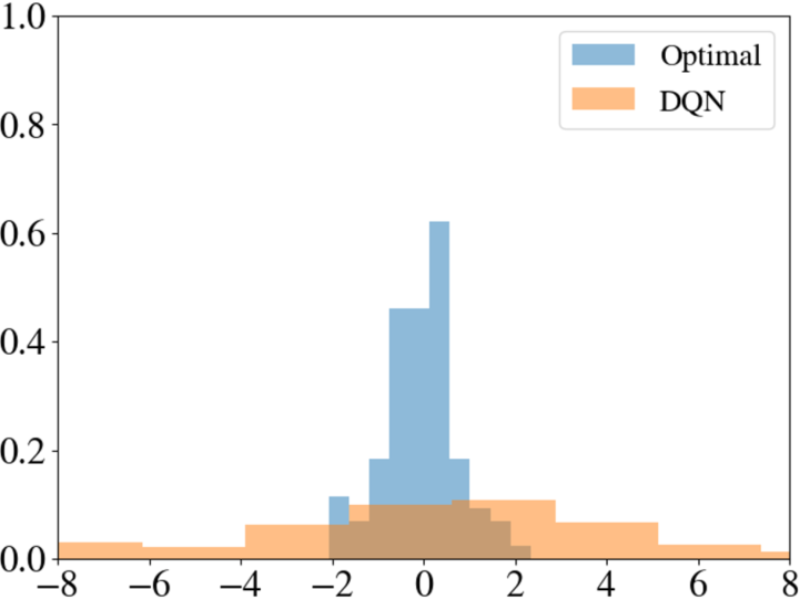}
		\caption{DQN - 20 samples per episode}
	\end{subfigure}
	\begin{subfigure}{0.45\linewidth}
		\includegraphics[width=\linewidth]{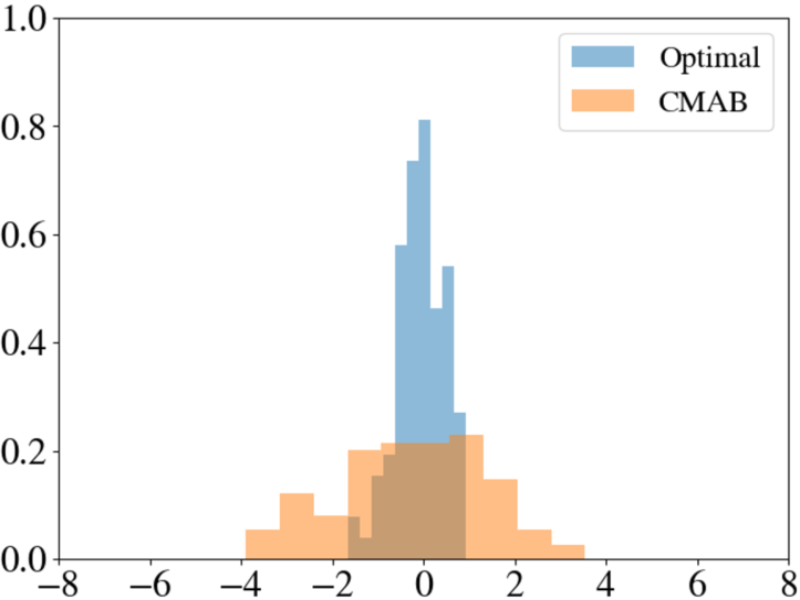}
		\caption{CMAB - 60 samples per episode}
	\end{subfigure}
	\hfil
	\rulesep
	\begin{subfigure}{0.45\linewidth}
		\includegraphics[width=\linewidth]{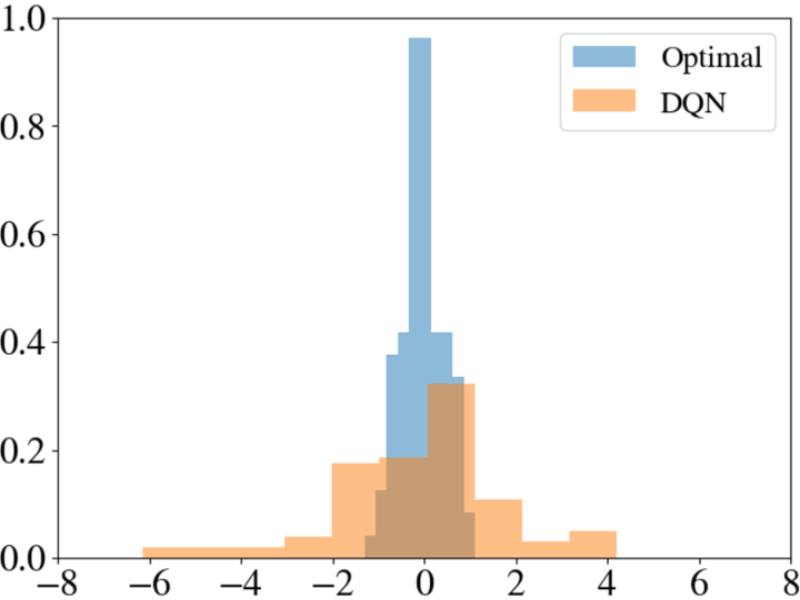}
		\caption{DQN - 60 samples per episode}
	\end{subfigure}
	\caption{Histograms of the terminal P\&L measured on $100$ test paths of CMAB, $\kappa=0$, (left column) and DQN (right column) agents in the non-deterministic scenario.}
	\label{plHists2}
\end{figure}

\begin{figure}[!htb]
	\centering
	\begin{subfigure}{0.45\linewidth}
		\includegraphics[width=\linewidth]{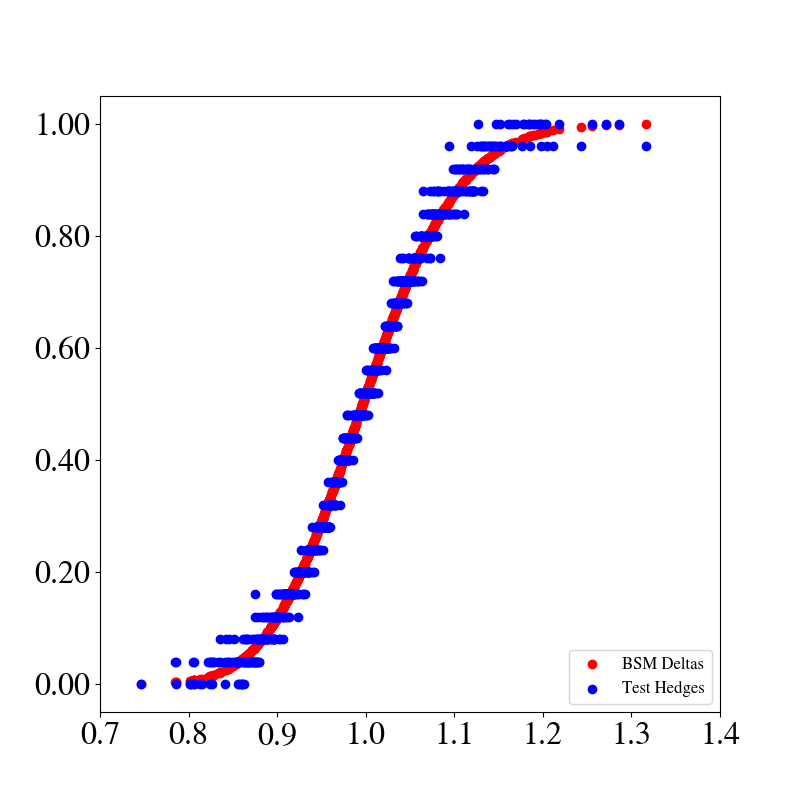}
		\caption{CMAB - 10 samples per episode}
	\end{subfigure}
	\hfil
	\rulesep
	\begin{subfigure}{0.45\linewidth}
		\includegraphics[width=\linewidth]{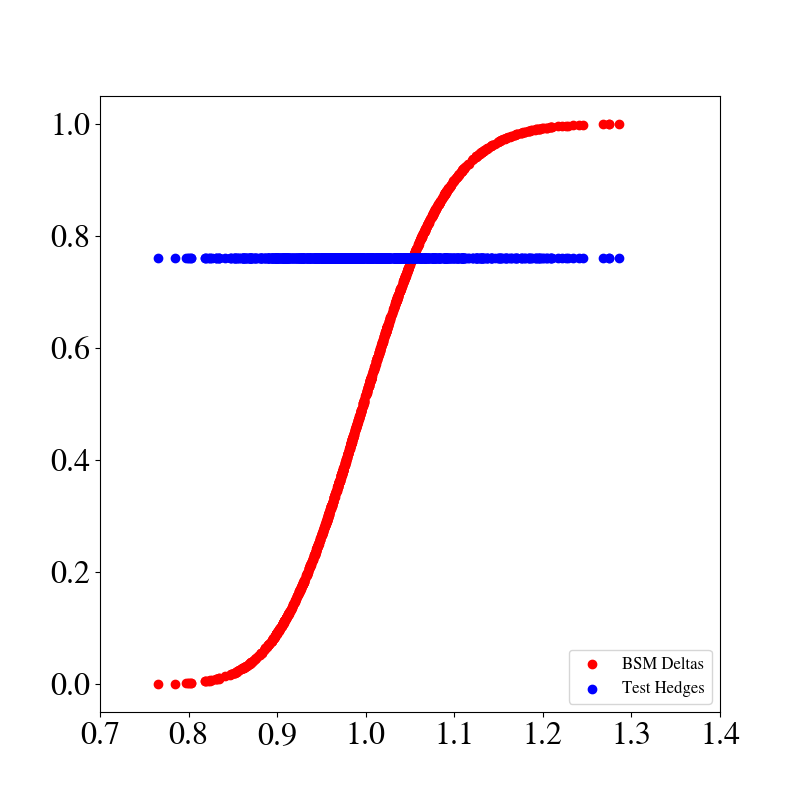}
		\caption{DQN - 10 samples per episode}
	\end{subfigure}
	
	\begin{subfigure}{0.45\linewidth}
		\includegraphics[width=\linewidth]{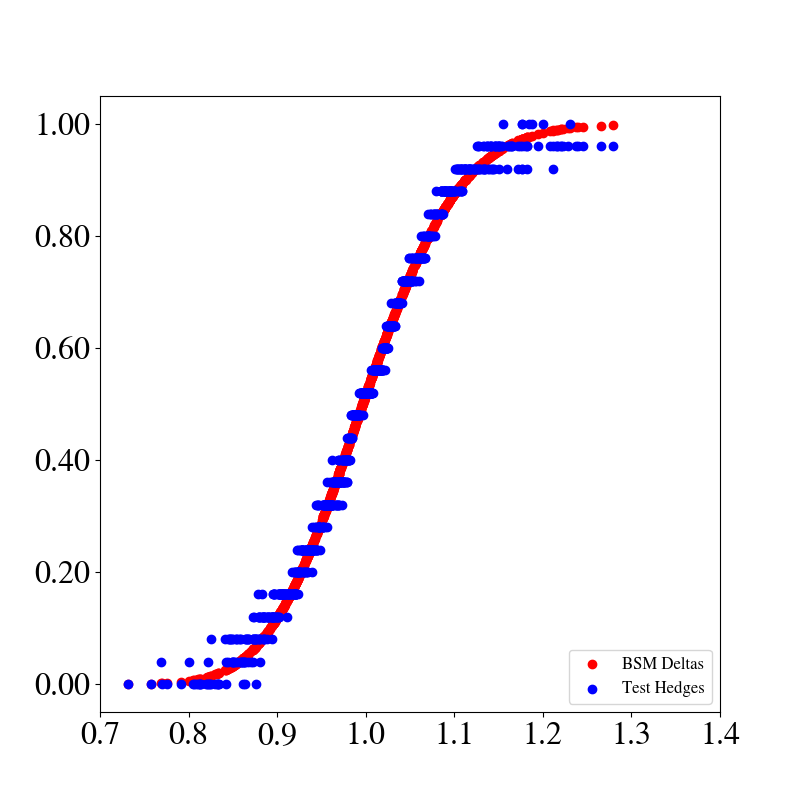}
		\caption{CMAB - 20 samples per episode}
	\end{subfigure}
	\hfil
	\rulesep
	\begin{subfigure}{0.45\linewidth}
		\includegraphics[width=\linewidth]{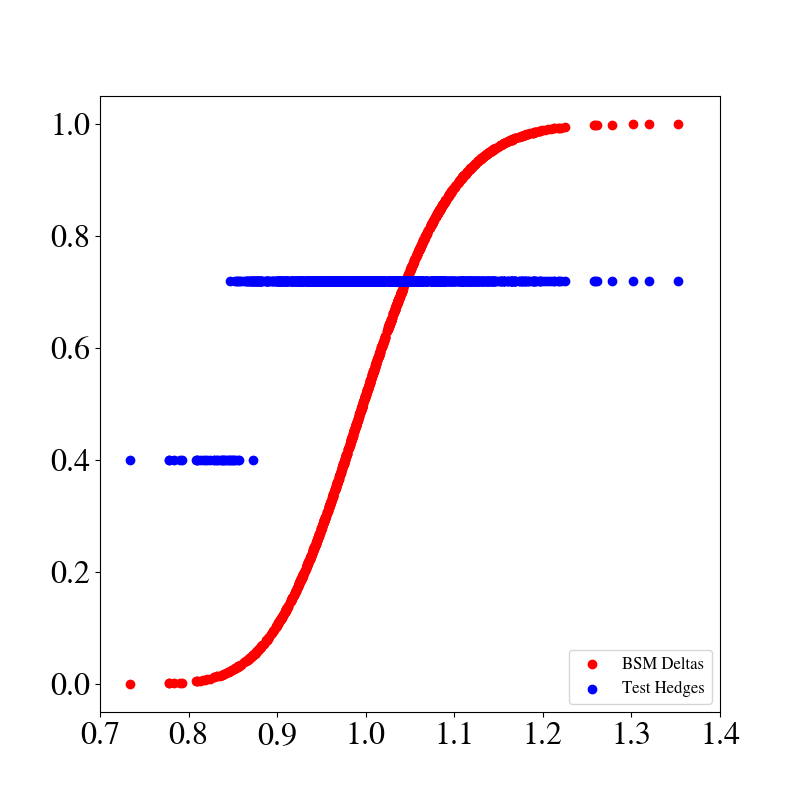}
		\caption{DQN - 20 samples per episode}
	\end{subfigure}
	
	\begin{subfigure}{0.45\linewidth}
		\includegraphics[width=\linewidth]{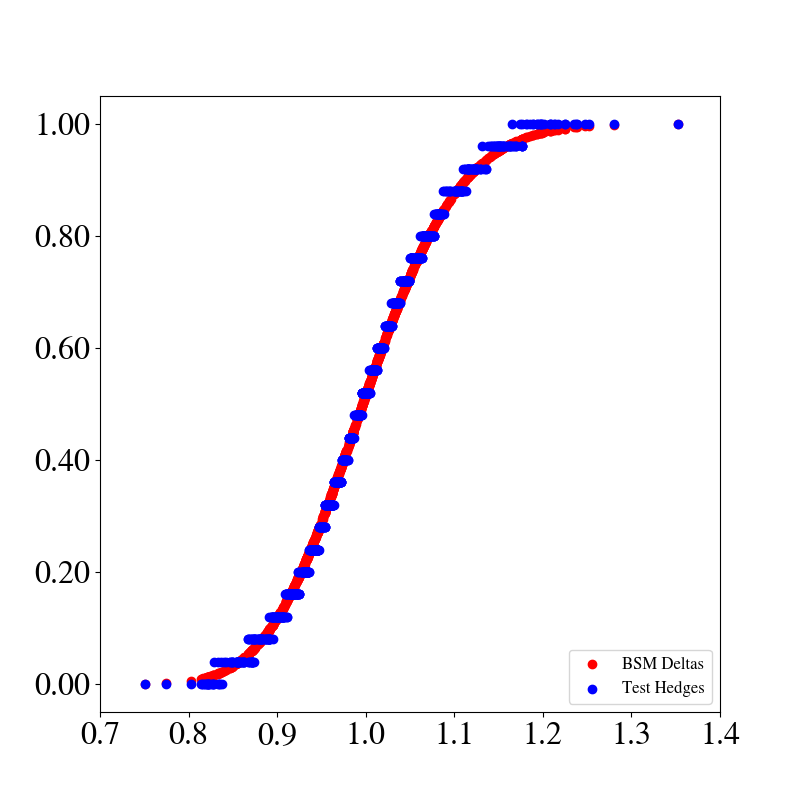}
		\caption{CMAB - 60 samples per episode}
	\end{subfigure}
	\hfil
	\rulesep
	\begin{subfigure}{0.45\linewidth}
		\includegraphics[width=\linewidth]{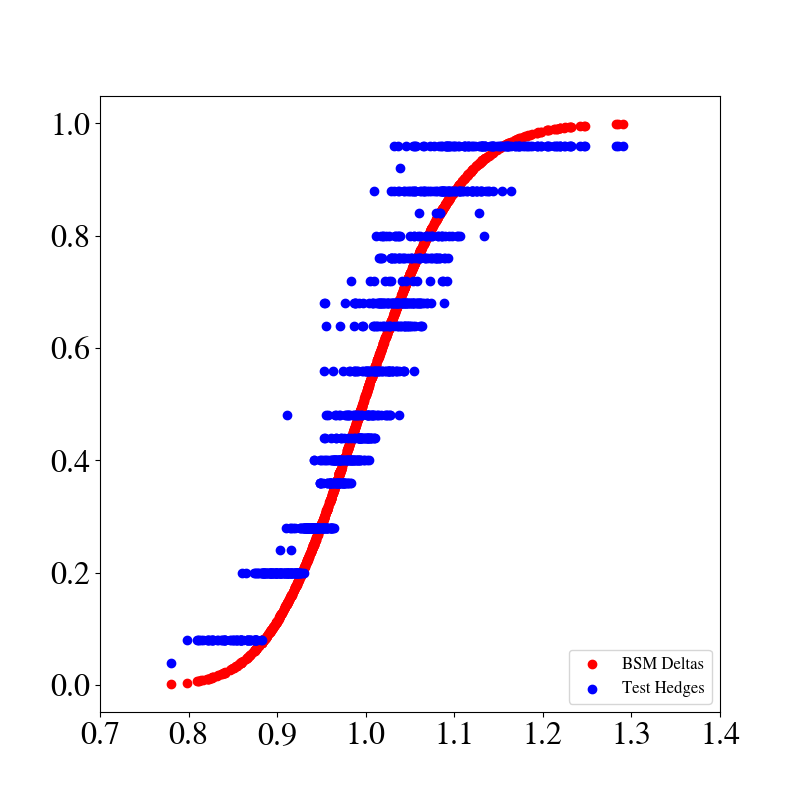}
		\caption{DQN - 60 samples per episode}
	\end{subfigure}
	\caption{Comparison of agent actions versus BSM $\delta$ at the mid-point of the investment horizon in deterministic scenario. $X$-axis: Value of underlying, $y$-axis: hedging action.}
	\label{deltasDeterministic}
\end{figure}

\begin{figure}[!htb]
	\centering
	\begin{subfigure}{0.45\linewidth}
		\includegraphics[width=\linewidth]{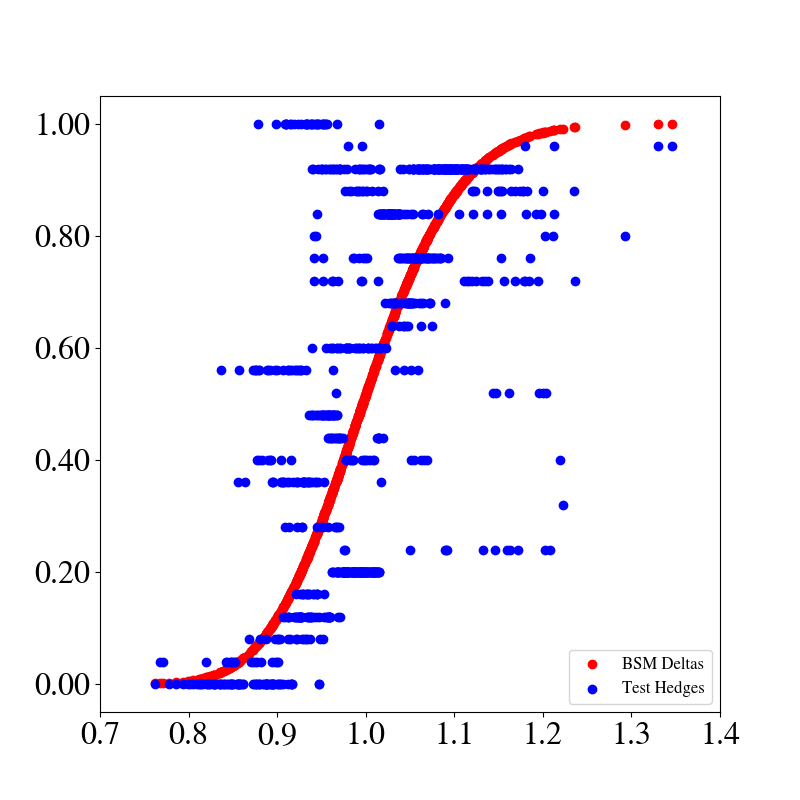}
		\caption{CMAB - 10 samples per episode}
	\end{subfigure}
	\hfil
	\rulesep
	\begin{subfigure}{0.45\linewidth}
		\includegraphics[width=\linewidth]{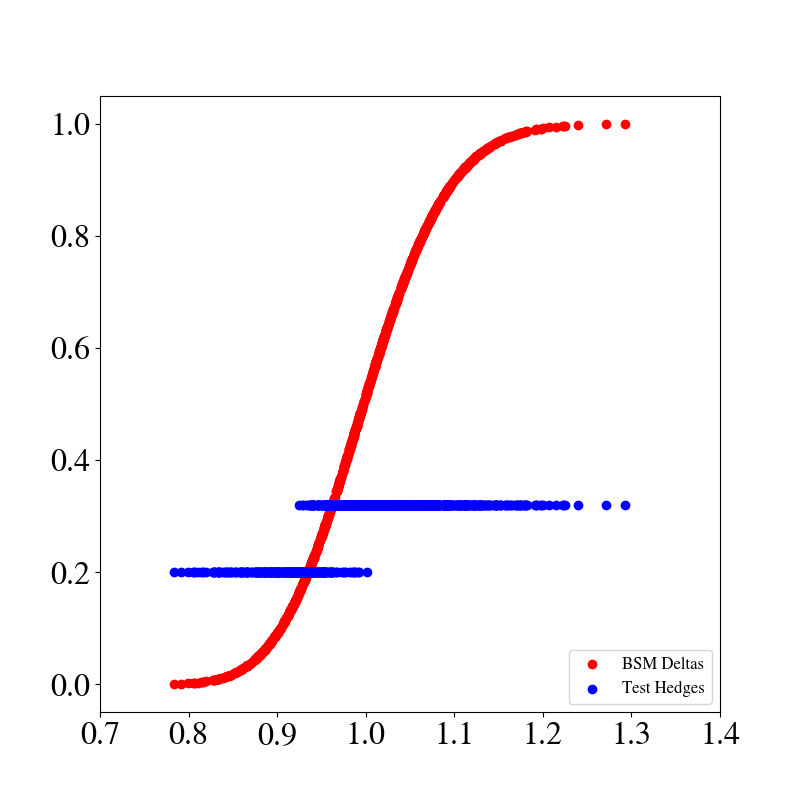}
		\caption{DQN - 10 samples per episode}
	\end{subfigure}
	
	\begin{subfigure}{0.45\linewidth}
		\includegraphics[width=\linewidth]{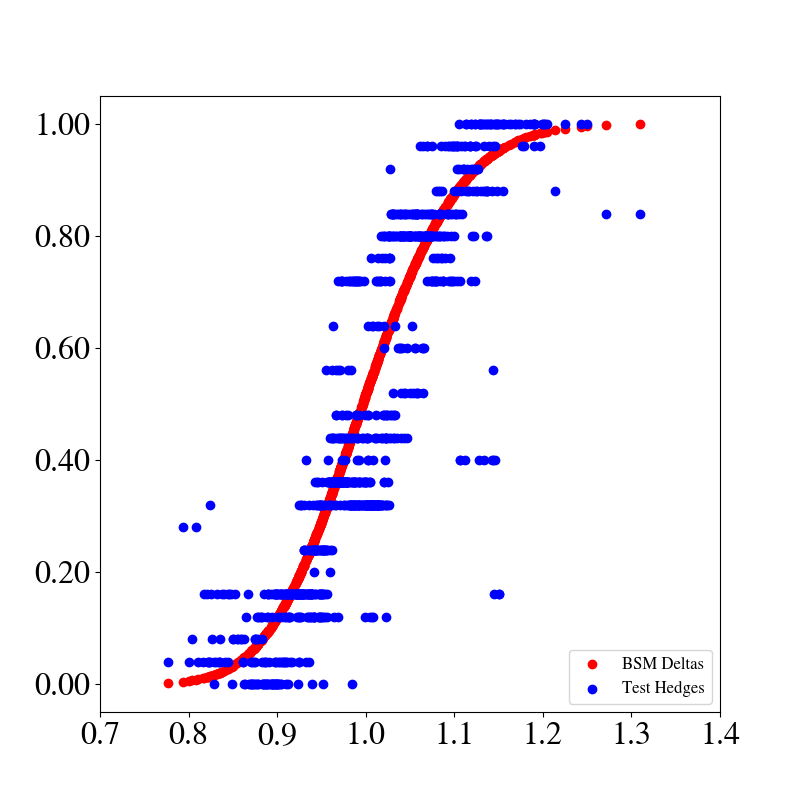}
		\caption{CMAB - 20 samples per episode}
	\end{subfigure}
	\hfil
	\rulesep
	\begin{subfigure}{0.45\linewidth}
		\includegraphics[width=\linewidth]{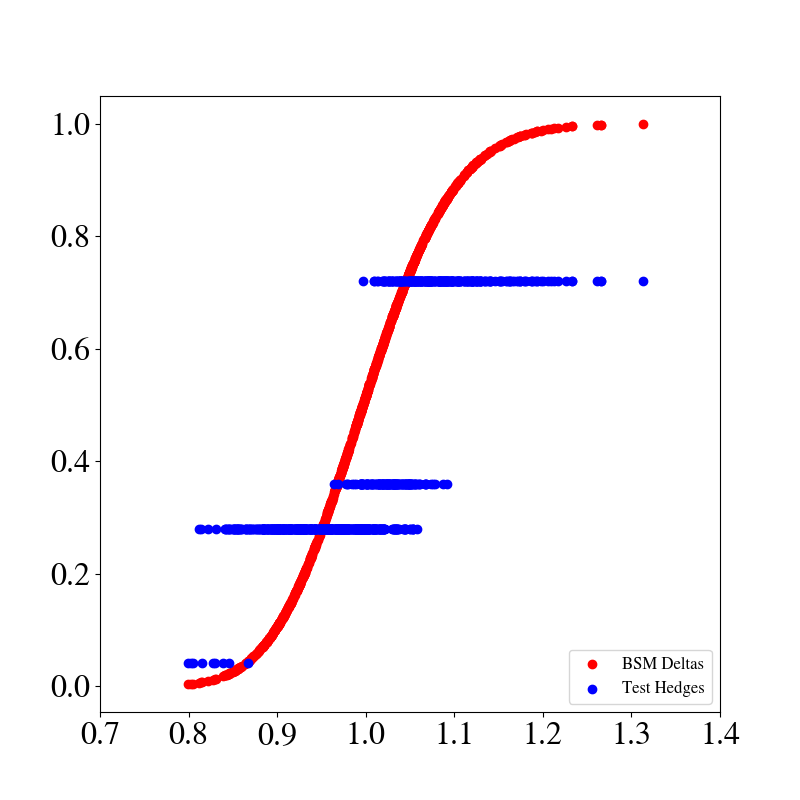}
		\caption{DQN - 20 samples per episode}
	\end{subfigure}
	
	\begin{subfigure}{0.45\linewidth}
		\includegraphics[width=\linewidth]{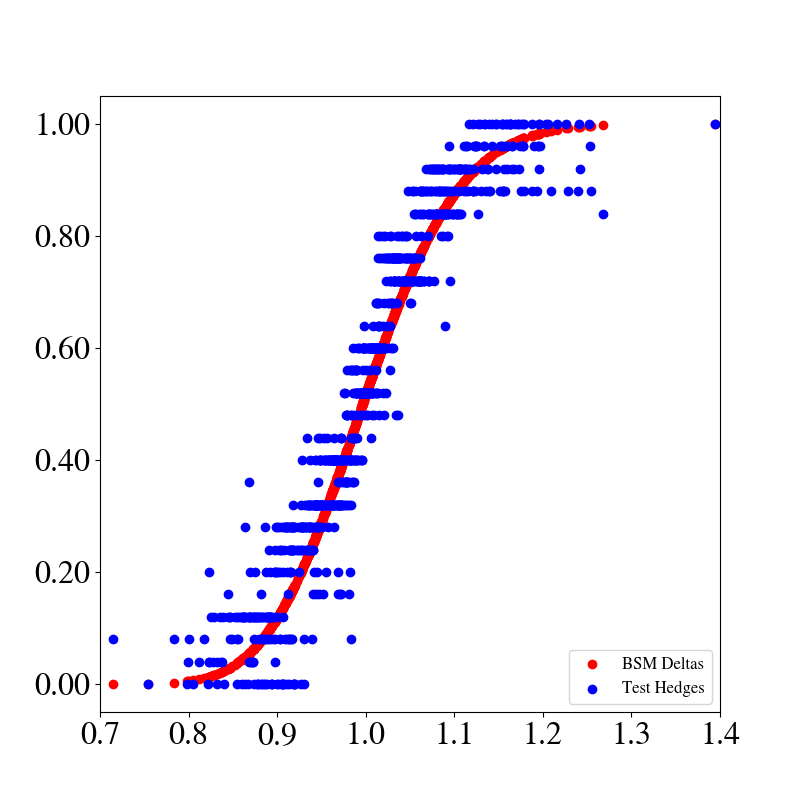}
		\caption{CMAB - 60 samples per episode}
	\end{subfigure}
	\hfil
	\rulesep
	\begin{subfigure}{0.45\linewidth}
		\includegraphics[width=\linewidth]{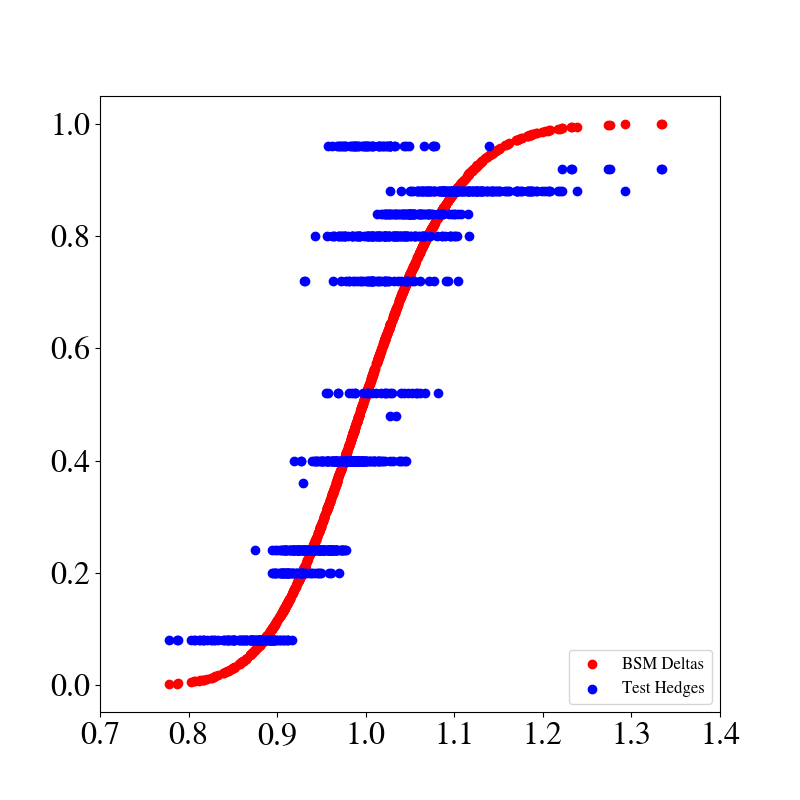}
		\caption{DQN - 60 samples per episode}
	\end{subfigure}
	\caption{Comparison of agent actions versus BSM $\delta$ at the mid-point of the investment horizon in non-deterministic scenario. $X$-axis: Value of underlying, $y$-axis: hedging action.}
	\label{deltasNonDeterministic}
\end{figure}

\begin{figure}[!ht]
	\centering
	\begin{subfigure}{\linewidth}
	    \centering
		\includegraphics[width=0.4\linewidth]{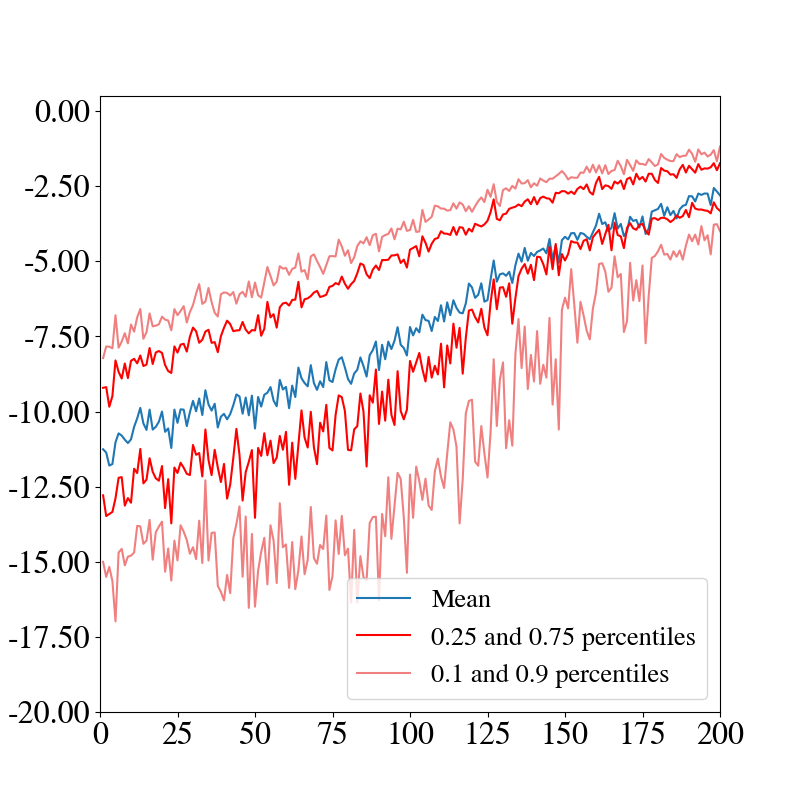}
		\includegraphics[width=0.45\linewidth]{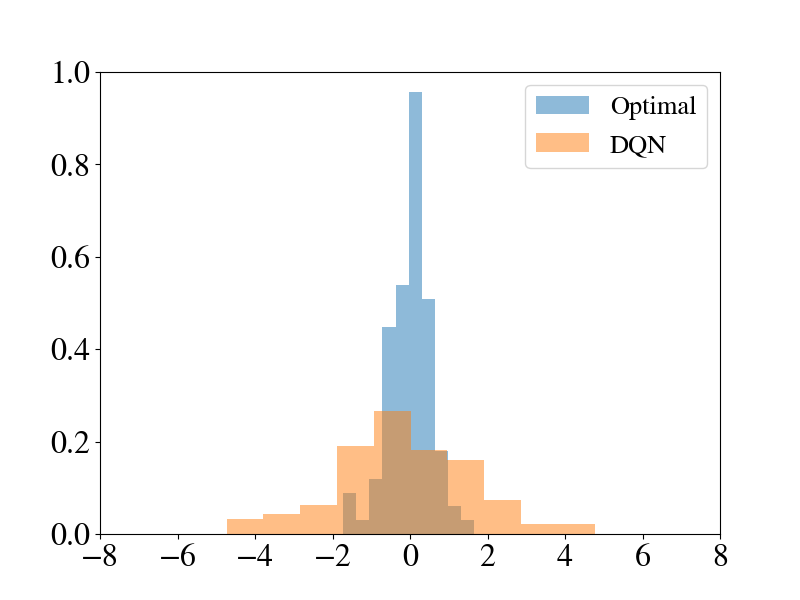}
		\caption{DQN - $\gamma=0.5$}
	\end{subfigure}
	
	\begin{subfigure}{\linewidth}
	    \centering
		\includegraphics[width=0.4\linewidth]{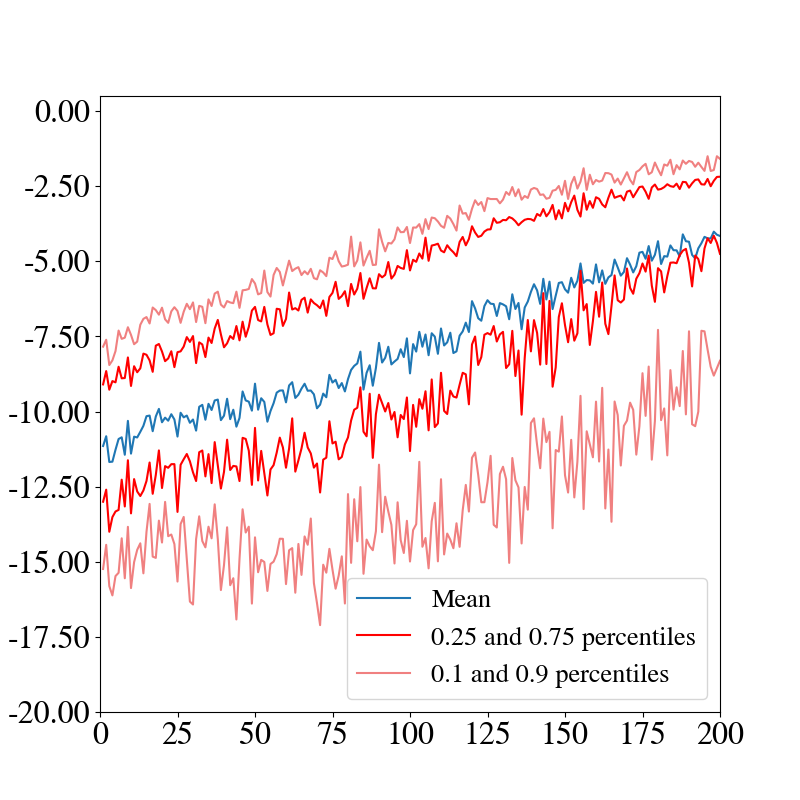}
		\includegraphics[width=0.45\linewidth]{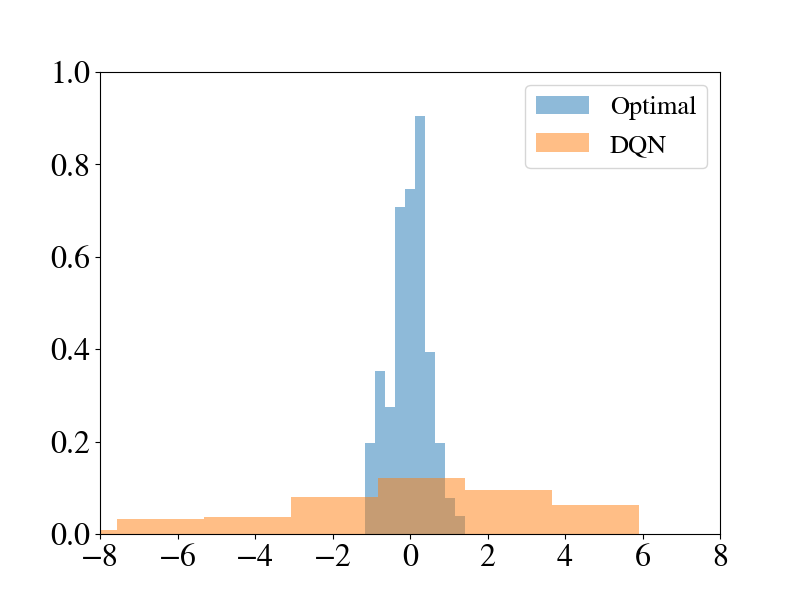}
		\caption{DQN - $\gamma=0.9$}
	\end{subfigure}
	
	\begin{subfigure}{\linewidth}
	    \centering
		\includegraphics[width=0.4\linewidth]{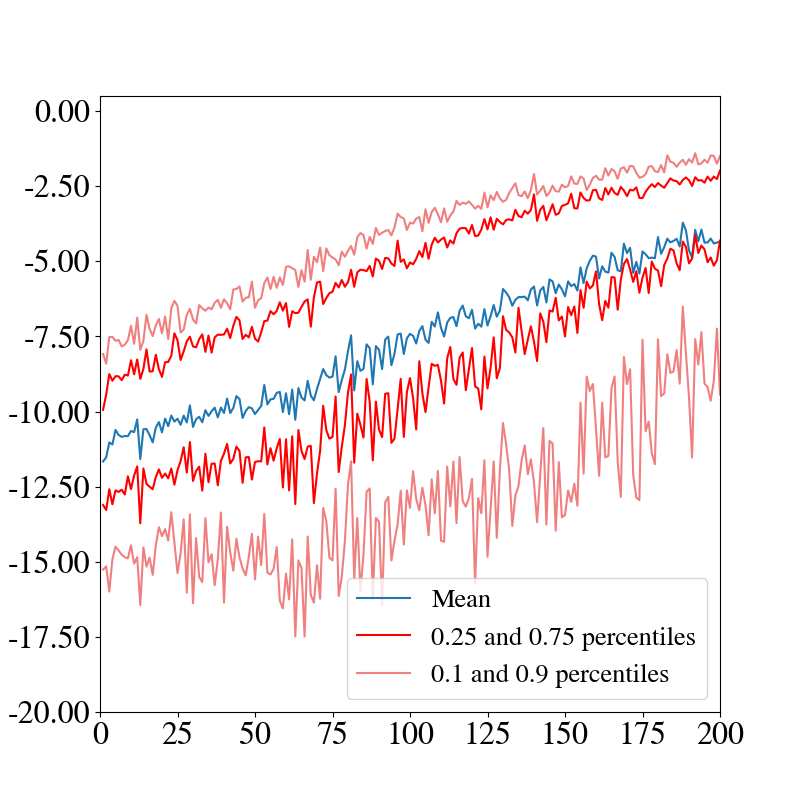}
		\includegraphics[width=0.45\linewidth]{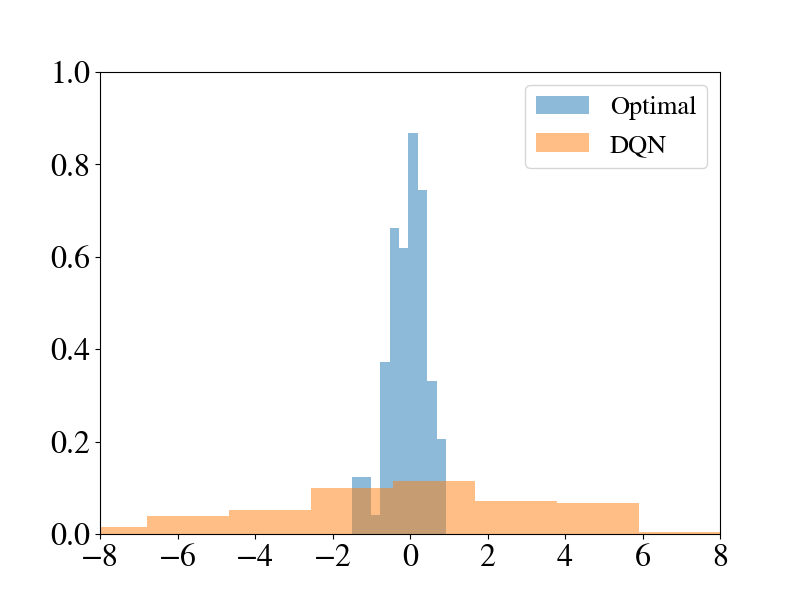}
		\caption{DQN - $\gamma=0.99$}
	\end{subfigure}
	
	\caption{Deterioration of training and test performance of DQN for P\&L hedging. Left column: Descriptive statistics over $100$ training cycles with $60$ samples per episode. Right column: Exemplary terminal P\&L histograms on $100$ test GBM paths.}
	\label{deteriorationDQN}
\end{figure}

As an illustration of the CMAB and DQN hedgers Figure~\ref{examplePaths} shows exemplary P\&L hedging paths obtained after training. To highlight the impact of discretization error, we also added the oracle agent, that knows the underlying market model and takes the optimal actions according to market expectations. We evaluate the hedging performance by measuring the terminal P\&L on $100$ GBM paths that share the same parameters but are not contained in the original training set. The resulting histograms of terminal P\&L are shown in Figure~\ref{plHists} for the deterministic and in Figure~\ref{plHists2} for the random market. To obtain a clearer picture about the actions chosen by the trained agents we compare them to the respective BSM $\delta$ on $1000$ test samples, where the impact of discretization becomes apparent. Figure~\ref{deltasDeterministic} and Figure~\ref{deltasNonDeterministic} show the dependency of the chosen actions on the underlying stock price. Finally Figure~\ref{deteriorationDQN} shows the dependency of DQN's performance on the choice of $\gamma$ in the non-deterministic scenario.

\subsection{Transaction Costs}\label{transactionCosts}
DQN is designed to address the full-fledged RL problem surpassing the planning capability of the CMAB algorithm. In this section we aim for evidence whether or not this capability aids in planning hedging decisions in the presence of transaction costs. For our experiment we consider proportional transaction costs of the form
\begin{align*}
cost(n_t,n_{t+1},S_t)=-\eta\cdot S_t\cdot\abs{n_{t+1}-n_t},
\end{align*}
where $n_t$ denotes the number of shares currently held, $n_{t+1}$ is the number of shares chosen for the coming period, and $\eta\geq0$ is a constant pre-factor. We choose $\eta=0.1$ for our experiments. The current holdings $n_t$ are passed to the agents as part of the context or the state. To asses whether the planning capability of DQN reduces transaction costs we choose a slightly different experimental setup for both agents. DQN is set up as a multi-period planning algorithm, which minimizes transaction costs over the entire lifetime of the option contract ($\gamma=0.9$). The current holdings $n_t$ are part of the state, both during training and testing. In contrast, the CMAB algorithm is trained to optimize immediate rewards by providing \emph{random} context about the number of shares. For testing, however, the choice $n_t$ of the previous period becomes the context of the next period. This corresponds to hedging in the real world, where the current holdings are a consequence of the preceding investment decision. 

\begin{figure}[!htb]
\centering
	\begin{subfigure}{0.45\linewidth}
		\includegraphics[width=\linewidth]{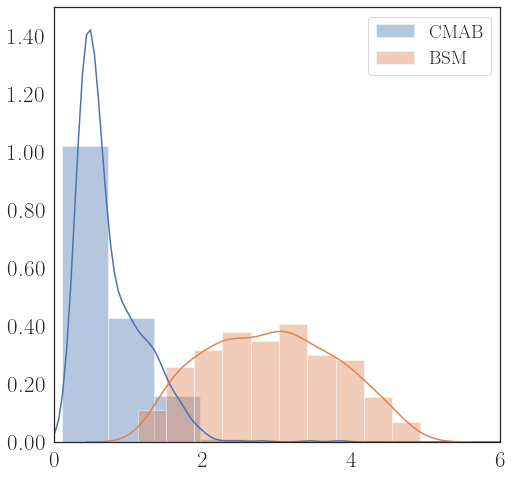}
		\caption{CMAB - transaction costs}
	\end{subfigure}
	\hfil
	\rulesep
	\begin{subfigure}{0.45\linewidth}
		\includegraphics[width=\linewidth]{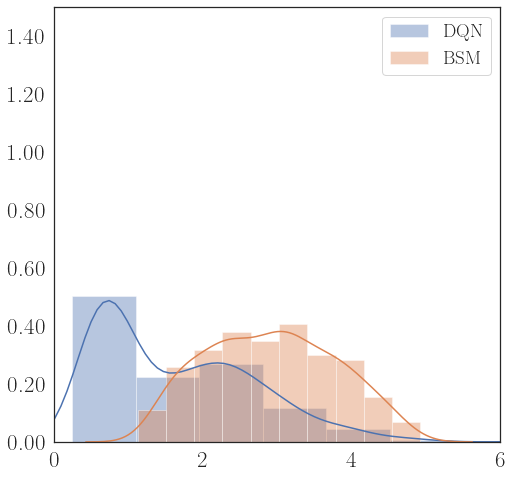}
		\caption{DQN - transaction costs}
	\end{subfigure}
	
\caption{Histograms of accumulated transaction costs measured on $1000$ testing episodes for CMAB (left) and DQN (right).}
\label{compareTransactionCosts}
\end{figure}

As before we grant $100$ training episodes to CMAB but $5000$ episodes to DQN. The accumulated transaction costs from a test set of $1000$ unseen episodes are compared in Figure~\ref{compareTransactionCosts}. As an immediate benchmark we also added the transaction costs that would emerge from hedging according to the BSM model. Notably, our transaction cost histogram for CMAB is comparable to the one obtained for DQN in~\cite[Exhibit 4]{kolm2019dynamic}. While~\cite{kolm2019dynamic} reports training on $5$ batches of $15000$ episodes with $50$ time steps each, we have used $100$ episodes with $60$ steps.
\subsection{Pre-Training and Adaptation}\label{preTraining}
\begin{figure}[!htb]
	\centering
	\begin{subfigure}{0.45\linewidth}
		\includegraphics[width=\linewidth]{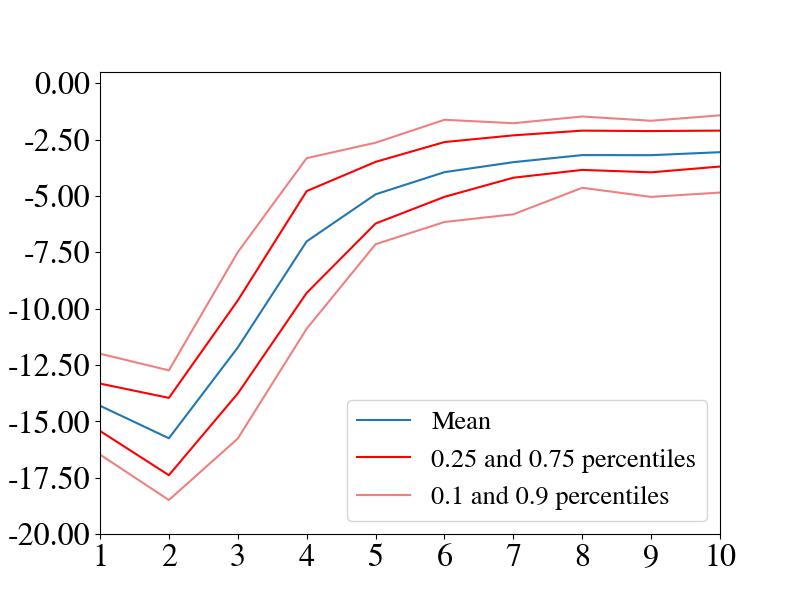}
		\caption{CMAB - Episodic rewards.}
	\end{subfigure}
	\rulesep
	\begin{subfigure}{0.45\linewidth}
		\includegraphics[width=\linewidth]{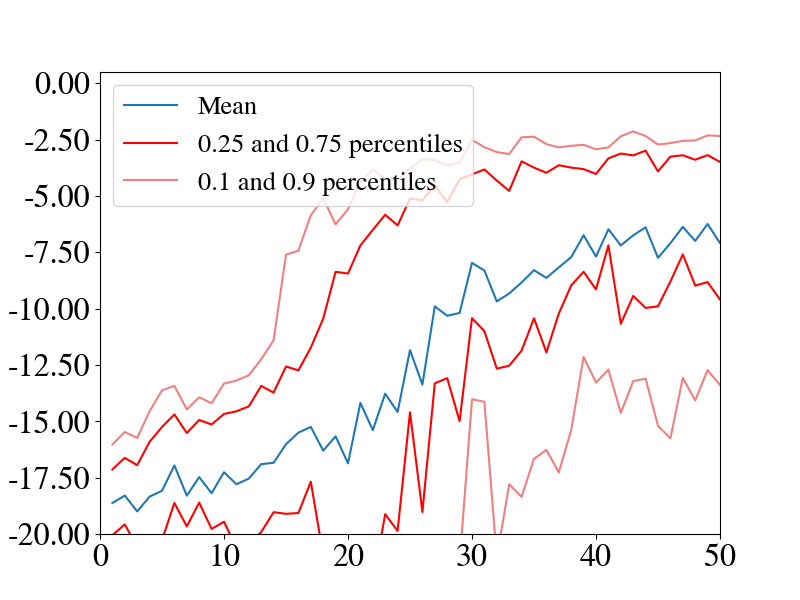}
		\caption{DQN - Episodic rewards.}
	\end{subfigure}
	\begin{subfigure}{0.45\linewidth}
		\includegraphics[width=\linewidth]{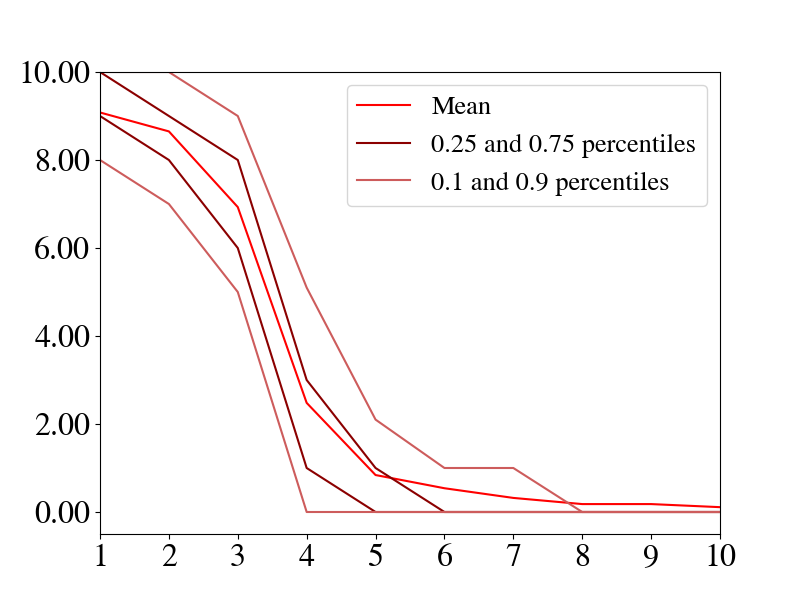}
		\caption{CMAB - Erroneous actions in the presence of market catastrophe, accumulated over training episodes.}
	\end{subfigure}
    \rulesep
	\begin{subfigure}{0.45\linewidth}
		\includegraphics[width=\linewidth]{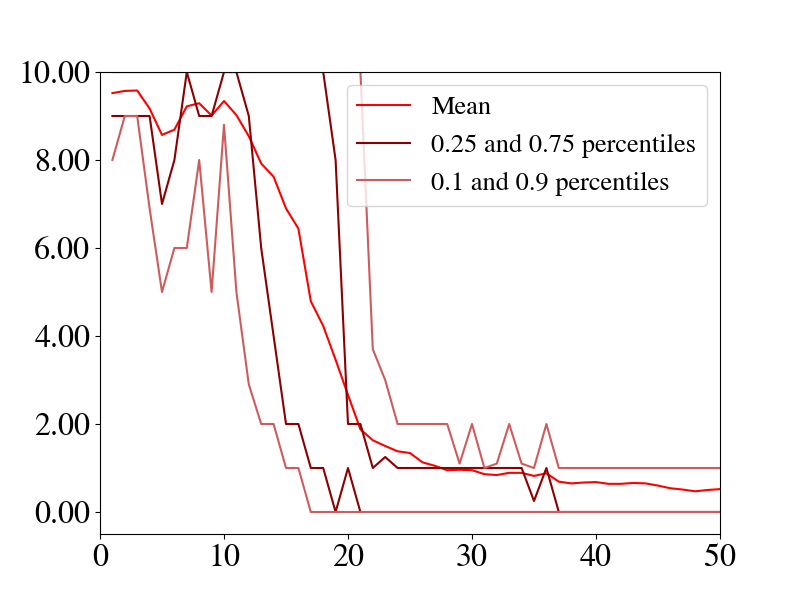}
		\caption{DQN - Erroneous actions in the presence of market catastrophe, accumulated over training episodes.}
	\end{subfigure}
	\caption{Adaptation of pre-trained CMAB (left column) and DQN (right column) agents to catastrophe context. Graphs show descriptive statistics from independent $100$ training cycles with $60$ samples per episode.}
	\label{trainingProgress_Catastrophe}
\end{figure}

\begin{figure}[!htb]
	\centering
	\begin{subfigure}{\linewidth}
	    \centering
		\includegraphics[width=0.45\linewidth]{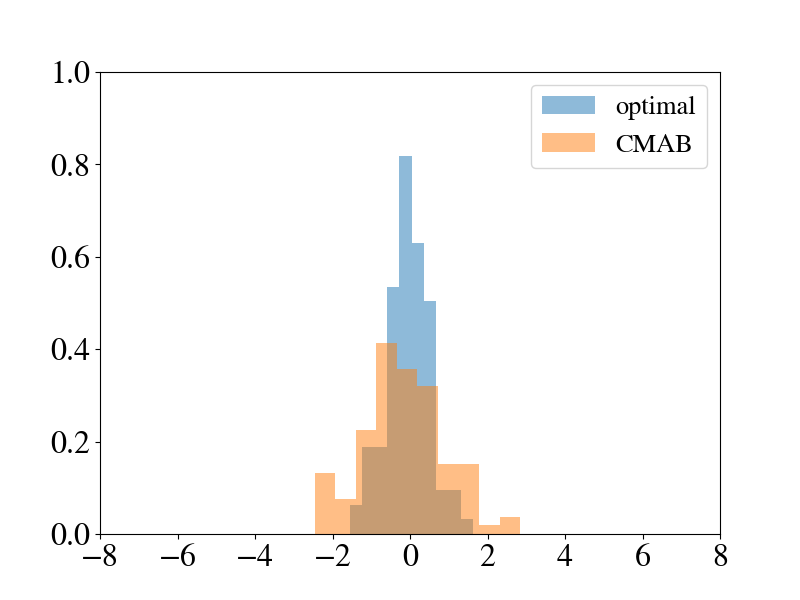}
		\includegraphics[width=0.45\linewidth]{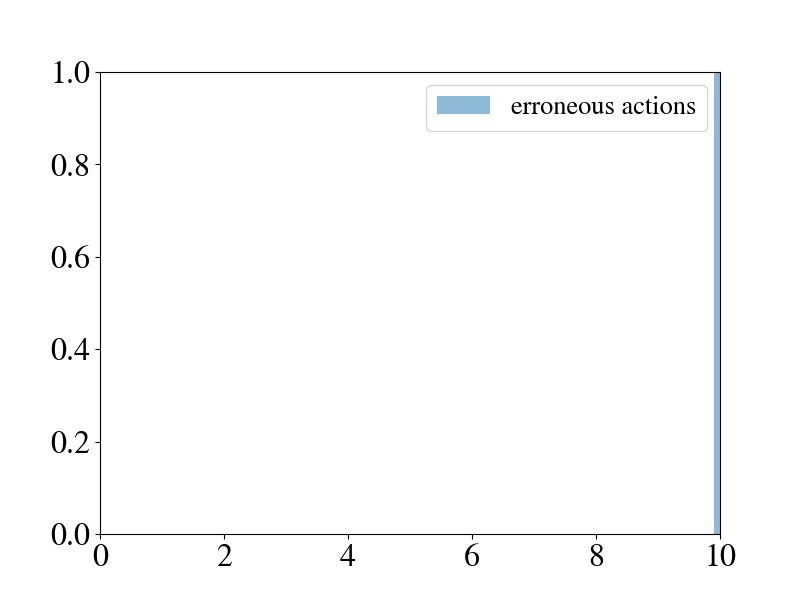}
		\caption{Pre-trained agent, before adaptation to market catastrophe context.}
	\end{subfigure}
	\begin{subfigure}{\linewidth}
	    \centering
		\includegraphics[width=0.45\linewidth]{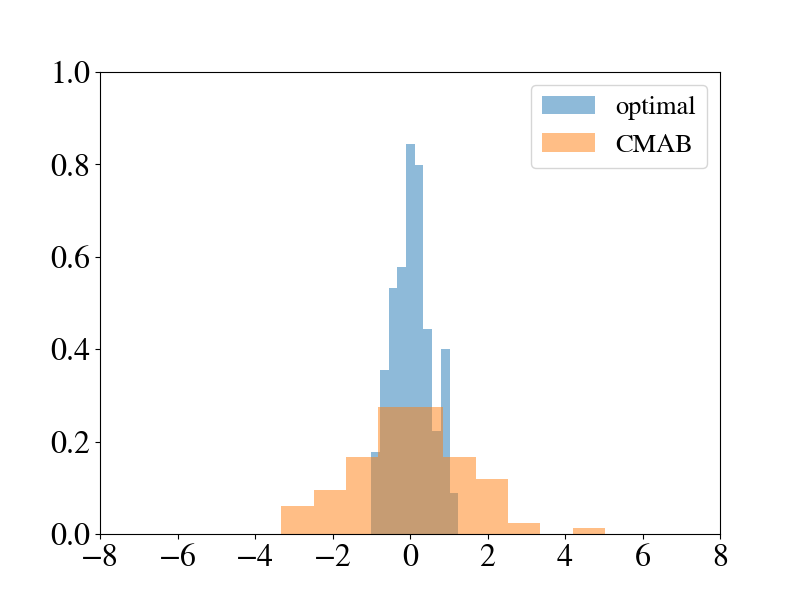}
		\includegraphics[width=0.45\linewidth]{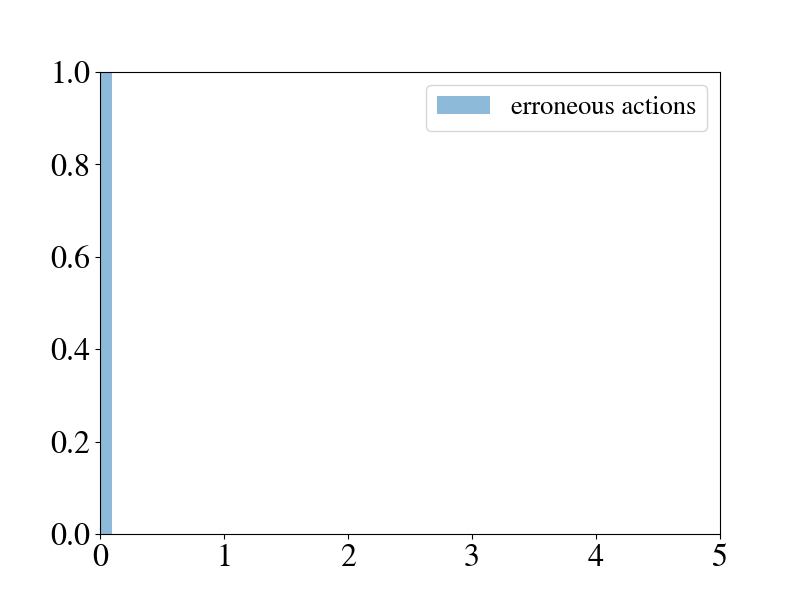}
		\caption{After adaptation to market catastrophe context.}
	\end{subfigure}
	\caption{Comparison of terminal P\&L (left column) and erroneous actions (right column) accumulated over test episodes for the CMAB agent.}
	\label{test_Catastrophe}
\end{figure}

In the practical work of an investment firm the availability of real-world data for training will be the bottle-neck for the deployment of any form of AI for hedging. We have seen in Section~\ref{trainingAlgos} that training from real-world market data (i.e.~in the non-deterministic scenario) is data costly, even for the minimalist' CMAB agent. A potential way out is that in applications the AI is \emph{pre-trained} on simulated data and then \emph{fine-tuned} on available real-world data. In this process sample-efficiency is key because only an efficient AI can adapt sufficiently fast to the new information. An inefficient AI, to the contrary, would simply over-fit to the chosen simulation models (for markets, transaction costs, etc...) and ignore the important real-world information. To analyse the adaptation capability of the CMAB and DQN algorithms we have designed the following experiment:

In the non-deterministic scenario the agent is pre-trained until it reaches the maximum reward threshold ($300$ episodes of $60$ samples for CMAB, $600$ episodes of $60$ samples for DQN, $\gamma=0.9$). This training process mimics the ``training on simulated data'' in practice. Once the model is fully trained we investigate how quickly it adapts to a previously unseen market reality. For this we invent random \emph{market catastrophe events}, which are provided as an additional binary context to the trained agent. In case of market catastrophe the agent should sell all holdings in the underlying immediately, otherwise it receives a large negative reward. In the absence of a catastrophe the agent should hedge as learned in pre-training. In the sequel we investigate the adaptation of the pre-trained agents to the new scenario. Since catastrophic events are rather sparse on the real markets we choose $10$ such events per training episode. An important point is that while catastrophes occur at the same time steps at each episode during training, in testing catastrophe events occur at random times. In other words we recycle the same sparse catastrophe data for training but we test in a more realistic setting with unpredictable times of catastrophe. We switched off short-selling for these experiments. The re-training progress of the pre-trained CMAB and DQN agents on episodes with added catastrophe events is depicted in Figure~\ref{trainingProgress_Catastrophe}.
The experimental findings demonstrate that after approximately $8$ training episodes the pre-trained CMAB agent has adapted to the new market context in terms of accumulated rewards and erroneous actions on catastrophe events. First CMAB agents show $0$ erroneous actions after $4$ training episodes. The DQN agent did not adapt fully to the new market setting even after $50$ training episodes. To gain a clearer picture on how well the CMAB agent adapted to the new context we compare the terminal P\&L statistics and the accumulated erroneous actions before and after re-training on $100$ previously unseen samples, see Figure~\ref{test_Catastrophe}. We observe that during the test cycle the trained agent did not choose a single erroneous action in case of market catastrophe, although such events occurred at random moments in time. The P\&L performance of the hedger has shown only slight deterioration. We conclude that the pre-trained CMAB agent successfully adapted to the new market information on a set of sparse market events (that have been used multiple times in the training cycle). This is a key feature of the CMAB agent as it is prerequisite to the deployment of pre-trained models in a practical setting.

\subsection{CMAB versus DQN for derivatives hedging in practice}\label{hedgingInPractice}

Conceptually our choice of a bandit-type formulation resembles business operations at an investment firm. It is rare that hedging decisions target the optimization of multi-period transaction costs and prices as in the full-fledged RL problem. Instead, traders consider the current market state and optimize for the immediate monetary targets within thresholds of immediate risk (which will almost certainly include a mark-to-market of the traded structure to an internally approved model). Transaction costs and risk quantifiers are monitored {on a daily basis} and reported to management and regulators. Most trading desks are subject to a forced liquidation upon reaching a maximum drawdown (or risk limit): this absorbing state has a profound impact on the theoretical value of an option, yet it is wholesale ignored in most modeling approaches. Long-term planning in the sense of taking sacrifice temporarily to achieve better terminal reward (as it is typical for games like Chess) is therefore often an impractical goal within realistic business constraints.

{Notably}, in the absence of transaction costs and risk-adjustment (and the limit of small time steps) the hedges obtained by the CMAB algorithm converge to the BSM $\delta$. This feature is desirable for benchmarking and not {present} in the $Q$-learning framework: While the value of an option contract in the BSM economy follows the BSM differential equation, $Q$-learning is designed to approximate solutions of the Bellman equation. As a consequence it is suitable for terminal utility hedging in the Cash Flow formulation, when the value function follows a Bellman equation, such as in~\cite{hodges1989option}. This discrepancy is also reflected in the optimal hedges:
in the classical BSM economy the optimal hedge is provided by the $\delta$-hedge portfolio, where $\delta$ only depends on the current price of the underlying and the remaining time to maturity. DQN anticipates the expected reward structure, a feature that fits naturally with the optimization of terminal utility, see Section~\ref{hedg}, but not the BSM {model}.

As a consequence the role of the parameter $\gamma$ in DQN (see Section~\ref{q_intro}) is important: apart from being a discount factor, it also regulates, in terms of the Bellman equation, the dependency between the immediate reward resulting from a given action and the future rewards. Experimental results are in line with the expectation that hedging {performance} become weaker (when benchmarked against BSM) when the DQN agent attempts to anticipate {the} future reward structure, see Figure~\ref{deteriorationDQN}. But what happens in the presence of propagating transaction costs? If costs are {{small}}, then probably, $\gamma$ should be comparatively {{small}}. To find the optimal value for $\gamma$ is a challenging task. In reality $\gamma$ {can} thus be viewed as a delicate hyperparameter that should be tuned to the~{real} markets. The right choice of risk-measure is not obvious either. Indeed, the preceding articles~\cite{hodges1989option,kolm2019dynamic,halperin2017qlbs,cao2019deep,ritter2017machine} rely on (multiple) {different} measures of risk. This implies that the same hedging path would be valued differently within the various setups and makes setting up a planning algorithm even more delicate. From a practical perspective, we remark that often DQN is tedious to train and unstable, especially when used in conjunction with NNs~\cite{tsitsiklis1997analysis}: We experimented with hedging in the Cash Flow formulation, where the frequent divergence of $Q$-values requires an architecture that stores snapshots of the algorithm state to allow restarting from the snapshot.

\section{Conclusions}\label{sec:conclude}
We have confirmed, following the preceding publications~\cite{hodges1989option,kolm2019dynamic,halperin2017qlbs,cao2019deep}, that RL agents can be trained to hedge derivative contracts. We have introduced the R-CMAB algorithm for P\&L hedging, which is motivated by its relative simplicity and sample efficiency. 

During training, RL agents learn to operate within their specific training environment. RL agents trained on simulated data perform well in market situations that are similar to the training simulations. This reveals an important shortcoming of RL methods when applied to hedging in practice: the lack of real-world data {leaves the practitioner with little options but} to train on simulated data. For full-fledged RL algorithms like DQN, the amount of simulated training data is huge as compared to the real-world data that the agent encounters in operation. The statistical influence of the real-world data will therefore often be negligible, leaving not much but an overfitting to the simulation models. We have demonstrated that the R-CMAB algorithm addresses this issue by a much more efficient and stable learning process. 

During execution, hedging agents should adapt to potential changes in the market. We have seen that the data cost of $Q$-learning will often render model updates unrealistic and the proposed CMAB algorithm reacts to new market conditions faster. In similar vein, although $Q$-learning is well-suited for multi-period hedging problems and for the computation of reservation and minimum risk prices, obtaining sufficient real-world data remains elusive. Given the constraints of investment firms the practical benefit of the theoretically more accurate prices is not obvious. 

Finally, more modern algorithms, such as neural MCTS, are available to address the multi-period problems more efficiently. The presence of hyperparameters, such as $\gamma$ in DQN, which regulates the inter-dependency of rewards and the level of risk-aversion, which tunes between different measures of risk, makes the evaluation of hedging performance subtle. {Indeed}, hedges obtained from R-CMAB naturally converge to BSM deltas when no risk adjustment, discretization error and transaction costs are present. This makes R-CMAB a useful benchmark for performance comparison and the development of more sophisticated algorithms. 

For completeness, we mention that direct minimization of terminal risk using Supervised Learning addresses the important use case of hedging complicated structures that cannot be priced with standard tools, but is much simpler than DQN.
While our findings are no surprise from the perspective of AI research, the question of training-data availability has not been discussed in detail in the hedging literature. To the contrary, we have observed a tendency towards more complex architectures in the recent publications, but data-availability might dictate a simple model. It is not uncommon that RL is used in Finance in situations, where large data sets are available for training. A typical example, where RL algorithms are used successfully is market-making, where trading occurs on a sub-microsecond time scale and produces an abundance of real-world data. The application of RL algorithms in a settings that provides relatively few data points, such as options hedging, remains an area of active research.

\clearpage
\bibliographystyle{plain}
\bibliography{Bibliography.bib}
\end{document}